%% file: main.tex
\newif\ificlrversion
\iclrversionfalse

\ificlrversion
    \documentclass{article}
    \usepackage{iclr2027_conference,times}
\else
    \documentclass[11pt]{article}
    \usepackage{Elastic_Robotics_Template/elasticreport}
\fi

\input{math_commands.tex}

\usepackage{hyperref}
\usepackage{url}
\usepackage{booktabs}
\usepackage{colortbl}
\definecolor{tablehead}{gray}{0.90}
\definecolor{tablesummary}{gray}{0.80}
\usepackage{subcaption}
\usepackage{cleveref}
\usepackage{array}
\usepackage{tabularx}
\usepackage{xltabular}
\usepackage{multirow}
\usepackage{amssymb}
\usepackage{fix-cm}
\usepackage{etoolbox}
\usepackage{placeins}

\newtoggle{final}
\toggletrue{final} % uncomment before submission

\usepackage[size=tiny]{todonotes}

\newcommand{\benchyes}{\textcolor{green!50!black}{$\checkmark$}}
\newcommand{\benchno}{\textcolor{red!75!black}{$\times$}}
\newcommand{\benchcount}[1]{\mbox{\textcolor{green!50!black}{#1}}}

\title{DexArena: A Unified Benchmark for Long-Horizon Dexterous Manipulation.}

\author{%
\parbox[t]{\dimexpr\textwidth-2\tabcolsep\relax}{%
\raggedright\bfseries
Haonan Ping\textsuperscript{5,*}, Jianshu Hu\textsuperscript{3,4,*}, Yimin Wang\textsuperscript{5}, Yi Zhu\textsuperscript{6}, \\
Pai Liang\textsuperscript{7}, Qiyan Jin\textsuperscript{8}, Qirui Bao\textsuperscript{1}, Ying Qian\textsuperscript{9}, \\
Xinyang Ding\textsuperscript{1}, Hongfei Qiao\textsuperscript{1}, Tianle Xue\textsuperscript{1}, Yucheng Bai\textsuperscript{1}, \\
Soo Wei Jie\textsuperscript{1}, Shuguang Li\textsuperscript{2,\ensuremath{\dagger}}, Yutong Ban\textsuperscript{1,3,4 \ensuremath{\dagger}} \\[0.5em]
\normalfont\small
\textsuperscript{1} Shanghai Jiao Tong University \\
\textsuperscript{2} Tsinghua University \\
\textsuperscript{3} Elastic Robotics \\
\textsuperscript{4} Shanghai Institute for Physical Artificial Intelligence and Robotics (PAIR) \\
\textsuperscript{5} University of Michigan -- Ann Arbor \\
\textsuperscript{6} National University of Singapore \\
\textsuperscript{7} University of Pennsylvania \\
\textsuperscript{8} Nanyang Technological University \\
\textsuperscript{9} Stanford University
}}

\ificlrversion
\else
\renewcommand{\ReportTitle}{DexArena: A Unified Benchmark for\\Long-Horizon Dexterous Manipulation}
\renewcommand{\ReportShortTitle}{DexArena}
\renewcommand{\ReportAuthors}{%
Haonan Ping\textsuperscript{5,*}, Jianshu Hu\textsuperscript{3,4,*}, Yimin Wang\textsuperscript{5}, Yi Zhu\textsuperscript{6},\\
Pai Liang\textsuperscript{7}, Qiyan Jin\textsuperscript{8}, Qirui Bao\textsuperscript{1}, Ying Qian\textsuperscript{9},\\
Xinyang Ding\textsuperscript{1}, Hongfei Qiao\textsuperscript{1}, Tianle Xue\textsuperscript{1}, Yucheng Bai\textsuperscript{1},\\
Soo Wei Jie\textsuperscript{1}, Shuguang Li\textsuperscript{2}, Yutong Ban\textsuperscript{1,3,4\ensuremath{\dagger}}}
\renewcommand{\ReportAffiliations}{%
\textsuperscript{1} Shanghai Jiao Tong University; \textsuperscript{2} Tsinghua University; \textsuperscript{3} Elastic Robotics\\
\textsuperscript{4} Shanghai Institute for Physical Artificial Intelligence and Robotics (PAIR)\\
\textsuperscript{5} University of Michigan -- Ann Arbor; \textsuperscript{6} National University of Singapore\\
\textsuperscript{7} University of Pennsylvania; \textsuperscript{8} Nanyang Technological University; \textsuperscript{9} Stanford University}
\renewcommand{\ReportContact}{%
\textsuperscript{*} Equal contribution. \textsuperscript{\ensuremath{\dagger}} Corresponding authors.\\
Haonan Ping (\texttt{phaonan@umich.edu}); Jianshu Hu (\texttt{jianshu-hu@sipair.org.cn});\\
Yutong Ban (\texttt{yban@sjtu.edu.cn})}
\renewcommand{\ReportLinks}{\href{https://sipair-elastic.github.io/DexArena-website/}{Project website}\quad /\quad\href{https://github.com/SIPAIR-Elastic/DexArena}{Code}\quad /\quad\href{https://huggingface.co/datasets/SIPAIR/DexArena}{Data}}
\renewcommand{\ReportKeywords}{Dexterous manipulation; long-horizon tasks; benchmarks; robot learning}
\hypersetup{pdftitle={DexArena: A Unified Benchmark for Long-Horizon Dexterous Manipulation},
    pdfauthor={Haonan Ping, Jianshu Hu, Yimin Wang, Yi Zhu, Pai Liang, Qiyan Jin, Qirui Bao, Ying Qian, Xinyang Ding, Hongfei Qiao, Tianle Xue, Yucheng Bai, Soo Wei Jie, Shuguang Li, Yutong Ban}}
\renewenvironment{abstract}{\begin{reportabstract}}{\end{reportabstract}}
\renewenvironment{table*}{\begin{table}}{\end{table}}
\fi

\ificlrversion
\fi
\begin{document}

\ificlrversion
    \maketitle
\else
    \MakeReportTitle
\fi

% Show ICLR author notes only when the title block is non-anonymous.
\ificlrversion
\ifbool{iclrfinal}{%
    \begingroup
    \renewcommand{\thefootnote}{}
    \footnotetext{%
        \textsuperscript{*} Equal contribution.
        \textsuperscript{\ensuremath{\dagger}} Corresponding authors.\\
        Contact following authors for questions about the paper:\\
        Haonan Ping (\texttt{phaonan@umich.edu}); Jianshu Hu (\texttt{jianshu-hu@sipair.org.cn});\\
        Shuguang Li (\texttt{lisglab@tsinghua.edu.cn}); Yutong Ban (\texttt{yban@sjtu.edu.cn}).%
    }
    \endgroup
}{}
\fi

\begin{abstract}
Long-horizon dexterous manipulation remains a fundamental challenge for robot learning, requiring the composition of multiple skills, precise bimanual coordination, generalizable visual understanding, and robust skill transitions.
Yet existing manipulation benchmarks largely emphasize isolated skills or support only a limited range of end-effectors, making these capabilities difficult to evaluate systematically.
We introduce \textbf{DexArena}, a benchmark designed to evaluate learning and generalization in long-horizon bimanual dexterous manipulation.
DexArena features \textbf{10 long-horizon tasks and 30 stage-level tasks} spanning diverse household and industrial scenarios.
Each long-horizon task is composed of explicitly defined stages with physically grounded success conditions, enabling reliable evaluation of both intermediate progress and end-to-end task completion.
A unified benchmark interface and verified human-teleoperated demonstrations enable testing different policy learning algorithms, with standardized simulation configurations, scene randomization, trajectory replay, and task-level evaluation.
Through extensive baseline experiments, we identify substantial challenges in precise manipulation, generalization to scene variation, and skill composition.
DexArena establishes a challenging and reproducible benchmark for developing robot learning methods that move toward generalizable long-horizon dexterous behavior.
\end{abstract}

\begin{figure}[h]
    \centering
    \includegraphics[width=\linewidth]{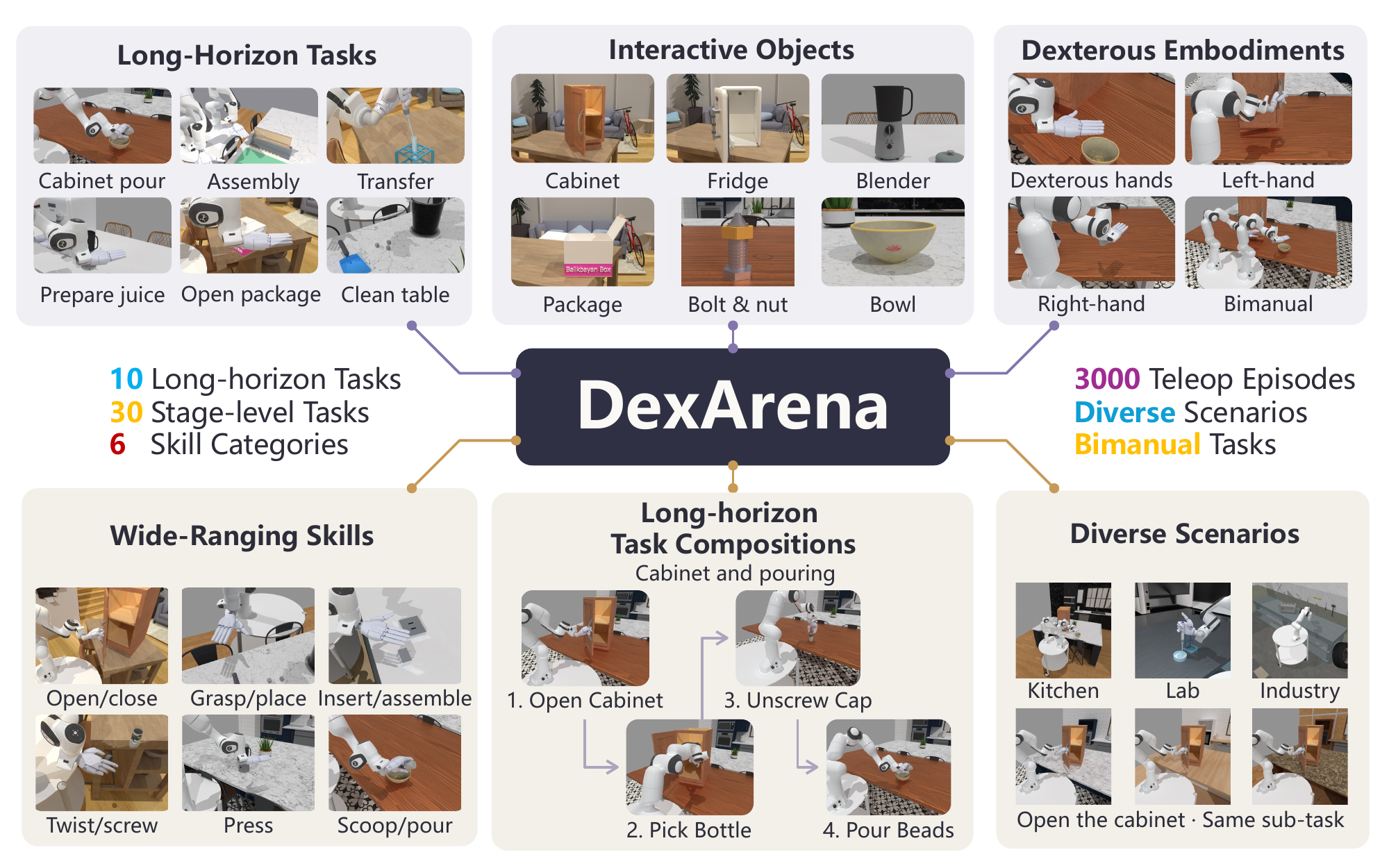}
    \caption{\textbf{DexArena overview.} A benchmark for long-horizon bimanual dexterous manipulation with multi-stage tasks, diverse scenarios, teleoperated demonstrations, and multiple embodiments.}
    \label{fig:teaser}
\end{figure}

\section{Introduction}
% \begin{enumerate}
%     \item Overview Figure
%     \item Comparison with other benchmarks (table)
%     \item Tasks Figure
%     \item Teleoperation System Figure
%     \item Data Statistics Figure 
%     \item Evaluation Results Table
% \end{enumerate}

Data-driven robot learning has made gripper-based manipulation increasingly capable.
By learning from expert demonstrations, imitation learning methods \citep{zhao2023act,chi2023diffusion} can acquire a wide range of manipulation skills, such as grasping, placing, and plugging.
However, much of this progress has focused on parallel-jaw end-effectors and short, isolated tasks.
Dexterous hands substantially expand the space of behaviors that robots can perform: multiple fingers can regrasp, twist objects, and manipulate tools.
These contact-rich, in-hand interactions are often beyond the capabilities of a parallel-jaw gripper.
At the same time, dexterous manipulation introduces a high-dimensional action space and complex contact dynamics, making learning robust policies for long-horizon tasks particularly challenging.

Recent progress in this area also requires an efficient and reproducible evaluation infrastructure.
Real-robot experiments are costly, time-consuming to reset, and difficult to reproduce across hardware platforms.
These challenges are especially pronounced for dexterous hands, whose complex mechanisms and precise actuation can limit the duration and scale of repeated experiments.
Physics-based simulators provide an attractive alternative by enabling controlled randomization, scalable interaction, and exact state reset.
However, beyond individual simulated tasks, there is a need for a benchmark that systematically evaluates whether a learned policy can execute dexterous skills, compose them across multiple dependent stages, and generalize across different scene configurations.

The existing benchmark landscape leaves this problem only partially addressed.
RLBench \citep{james2019rlbench}, Meta-World \citep{yu2020metaworld}, ManiSkill2 \citep{gu2023maniskill2}, RoboCasa \citep{nasiriany2024robocasa}, and RoboTwin \citep{robotwin2025} provide broad coverage of manipulation skills, household scenarios, or dual-arm settings, but primarily emphasize gripper-based manipulation.
Dexterous benchmarks, including Adroit \citep{rajeswaran2018learningcomplexdexterousmanipulation}, DexArt \citep{Bao_2023_CVPR}, DexMimicGen \citep{jiang2025dexmimicgenautomateddatageneration}, and DexJoCo \citep{dexjoco2026}, have begun to address dexterous hand manipulation, even for bimanual tasks.
However, long-horizon composition across dependent stages is rarely evaluated under a shared task definition.
Consequently, it remains difficult to disentangle whether a method has learned a dexterous skill itself, can reliably compose skills over long horizons, or is specialized to a particular visual configuration.

We introduce DexArena, a unified benchmark designed to make these capabilities measurable within a common physical framework.
DexArena contains 10 multi-stage tasks and 30 stage-level tasks spanning household and industrial-style scenarios.
Each multi-stage task consists of two to five stages, while each stage corresponds to an exact slice of its parent sequence: the multi-stage task and its stages share the same workstation, dynamics, and randomization ranges, with identical success conditions for each corresponding stage.
This design enables controlled evaluation of isolated skill execution and skill composition without changing the underlying physical setting.
Task completion is determined by explicit criteria, including articulated-object angles, relative poses, and insertion depths.
Success must persist for a specified duration, preventing transient contacts or unstable configurations from being counted as successful completion.

DexArena further separates skill learning from embodiment and rendering choices.
The benchmark provides a common interface across a parallel-jaw gripper and three dexterous hands (Sharpa Hand, Wuji Hand, and XHand), and supports both single-arm and bimanual manipulation.
All environments use standardized physical parameterizations and workstation configurations. 
Every recorded frame stores the scene state, so an episode can be restored frame by frame for re-rendering, and its recorded actions can be replayed from the recorded initial state.
Importantly, recorded trajectories can also be rendered under different camera configurations and background scenes after collection, enabling controlled evaluation of visual generalization without changing the underlying physical interaction.
To support data-driven learning, we collect human-teleoperated demonstrations for every stage-level task, providing physically valid trajectories for later evaluations. 

Using this infrastructure, we evaluate three representative policy families: demonstration-guided reinforcement learning, single-task imitation learning, and Vision-Language-Action (VLA) models.
We measure stage-level task success, generalization to scene variations, and end-to-end completion under sequential stage execution.
These evaluations expose different failure modes, distinguishing failures in contact-rich skill execution, visual generalization, and long-horizon skill composition.
Together, DexArena provides a common empirical framework for studying the transition from isolated manipulation skills to robust, compositional, and long-horizon dexterous behavior.

We summarize our contributions as follows:
\begin{enumerate}
\item We introduce DexArena, a unified benchmark for long-horizon, bimanual dexterous manipulation, with 10 multi-stage tasks and 30 physically grounded stage-level tasks that support evaluation of skill execution, generalization and composition.

\item We provide a reproducible benchmark infrastructure with four heterogeneous end-effectors, standardized simulation and randomization configurations, trajectory replay, and stage-level/task-level success criteria.
In particular, we invest substantial effort in collecting and validating physically valid, replayable demonstrations, enabling consistent reproduction of the demonstrated interactions and providing high-quality data for policy learning.

\item We establish representative baselines spanning imitation learning, demonstration-guided reinforcement learning, and vision-language-action models, and evaluate them on stage-level task performance, scenario generalization, and long-horizon composition to provide a common reference point for future methods.

\end{enumerate}

\section{Related Work}
Most existing manipulation benchmarks focus on gripper-based manipulation and have established the foundations for systematic evaluation of robot manipulation.
RLBench ~\citep{james2019rlbench} provides procedurally generated vision-based tasks and demonstrations.
Meta-World ~\citep{yu2020metaworld} evaluates multi-task and meta-reinforcement learning across a family of tabletop skills. 
ManiSkill2 ~\citep{gu2023maniskill2} provides GPU-parallel simulation, diverse assets, and standardized interfaces for large-scale manipulation experiments.
RoboCasa ~\citep{nasiriany2024robocasa} extends evaluation to household manipulation through procedurally generated scenes and language-conditioned tasks. 
More recent benchmarks, including RoboTwin~\citep{robotwin2025} and RoboDojo~\citep{robodojo2026}, further explore scalable data generation and evaluation for bimanual and gripper-based manipulation.
Together, these benchmarks demonstrate the value of standardized simulation for studying visual generalization, multi-task learning, and manipulation policy performance.

Dexterous hands introduce additional challenges including underactuated or high-dimensional control, contact-rich interactions, and coordination across multiple fingers.
Adroit ~\citep{rajeswaran2018learningcomplexdexterousmanipulation} studies object relocation, in-hand manipulation, tool use, and door opening with a dexterous hand, using human demonstrations to support policy learning.
DexArt ~\citep{Bao_2023_CVPR} focuses on articulated-object manipulation with dexterous hands.
DexMimicGen ~\citep{jiang2025dexmimicgenautomateddatageneration} presents nine bimanual dexterous-manipulation environments across three embodiments and synthesizes demonstrations from a small set of human teleoperation demonstrations.
DexJoCo ~\citep{dexjoco2026} studies 11 dexterous manipulation tasks based on MuJoCo simulation \citep{todorov2012mujoco}.
These efforts motivate benchmark designs that evaluate more dexterous manipulation tasks and consider the ability to compose skills.

Despite this progress, existing benchmarks provide different combinations of task diversity, horizon, embodiment coverage, and evaluation granularity.
DexArena complements these efforts by providing a unified benchmark for long-horizon dexterous manipulation across heterogeneous end-effectors.
It features explicitly staged tasks with physically grounded success conditions and supports evaluation of both intermediate progress and full-task completion.
By combining multi-stage task composition and embodiment variation within a common simulation framework, DexArena enables systematic evaluation of the capabilities required for complex dexterous manipulation.

\begin{table}[t]
\centering
\caption{\textbf{Capability comparison with related benchmarks.}
Green checks indicate emphasized capabilities.
Red crosses indicate capabilities not emphasized.}
\label{tab:comparison_with_other_benchmarks}
\scriptsize
\setlength{\tabcolsep}{1pt}
\renewcommand{\arraystretch}{1.2}
\renewcommand{\tabularxcolumn}[1]{m{#1}}
\begin{tabularx}{\linewidth}{@{}l*{7}{>{\centering\arraybackslash}X}@{}}
\toprule
\rowcolor{tablehead}
\textbf{Benchmark} & \shortstack{Num. of\\tasks} & \shortstack{Dexterous\\hands} & \shortstack{Bimanual\\tasks} & \shortstack{Cross-\\embodiment} & \shortstack{Teleop\\demos} & \shortstack{Long-\\horizon} & \shortstack{Task\\composition} \\
\midrule
Adroit~\citep{rajeswaran2018learningcomplexdexterousmanipulation} & 4 & \benchyes & \benchno & \benchno & \benchcount{100} & \benchno & \benchno \\
RLBench~\citep{james2019rlbench} & 100 & \benchno & \benchno & \benchno & \benchno & \benchno & \benchno \\
Meta-World~\citep{yu2020metaworld} & 50 & \benchno & \benchno & \benchno & \benchno & \benchno & \benchno \\
ManiSkill2~\citep{gu2023maniskill2} & 20 & \benchno & \benchno & \benchno & \benchno & \benchno & \benchno \\
DexArt~\citep{Bao_2023_CVPR} & 4 & \benchyes & \benchno & \benchno & \benchno & \benchno & \benchno \\
RoboCasa~\citep{nasiriany2024robocasa} & 100 & \benchno & \benchno & \benchno & \benchno & \benchyes & \benchyes \\
DexMimicGen~\citep{jiang2025dexmimicgenautomateddatageneration} & 9 & \benchyes & \benchyes & \benchyes & \benchcount{60} & \benchyes & \benchno \\
RoboTwin2.0~\citep{robotwin2025} & 50 & \benchno & \benchyes & \benchyes & \benchno & \benchno & \benchno \\
RoboDojo~\citep{robodojo2026} & 60 & \benchno & \benchyes & \benchyes & \benchcount{3500} & \benchyes & \benchno \\
DexJoCo~\citep{dexjoco2026} & 11 & \benchyes & \benchyes & \benchyes & \benchcount{1100} & \benchyes & \benchno \\
\rowcolor{tablesummary}
\textbf{DexArena (ours)} & 10+30 & \benchyes & \benchyes & \benchyes & \benchcount{3,000} & \benchyes & \benchyes \\
\bottomrule
\end{tabularx}
\end{table}

Due to the page limit, additional related work on policy learning is included in \Cref{subsec:related_work_policy_learning}.

\section{DexArena Benchmark}
\textbf{DexArena} is a physics-based simulation benchmark for long-horizon dexterous manipulation built on the SAPIEN simulator \citep{Xiang_2020_CVPR}.
This simulator allows us to run multiple environments in parallel for fast inference while maintaining high fidelity scene rendering.
DexArena comprises \textbf{10 multi-stage manipulation tasks and 30 stage-level tasks}, spanning scenarios from household activities to industrial-style manipulation, with both single-arm and bimanual settings.
Each multi-stage task is decomposed into explicitly defined stages with physically grounded success conditions, enabling a policy to be evaluated both on individual skills and on their composition into a complete task.
The benchmark supports four end-effector types: a parallel-jaw gripper and three dexterous hands (Sharpa Hand, Wuji Hand, and XHand).
By combining long-horizon task composition, contact-rich dexterous interaction, multiple embodiments, and standardized evaluation, DexArena provides a unified testbed for studying the capabilities and limitations of modern robot learning methods.

\subsection{Task Design}
\label{subsec:task_design}
\begin{figure}[t]
    \centering
    \includegraphics[width=\linewidth]{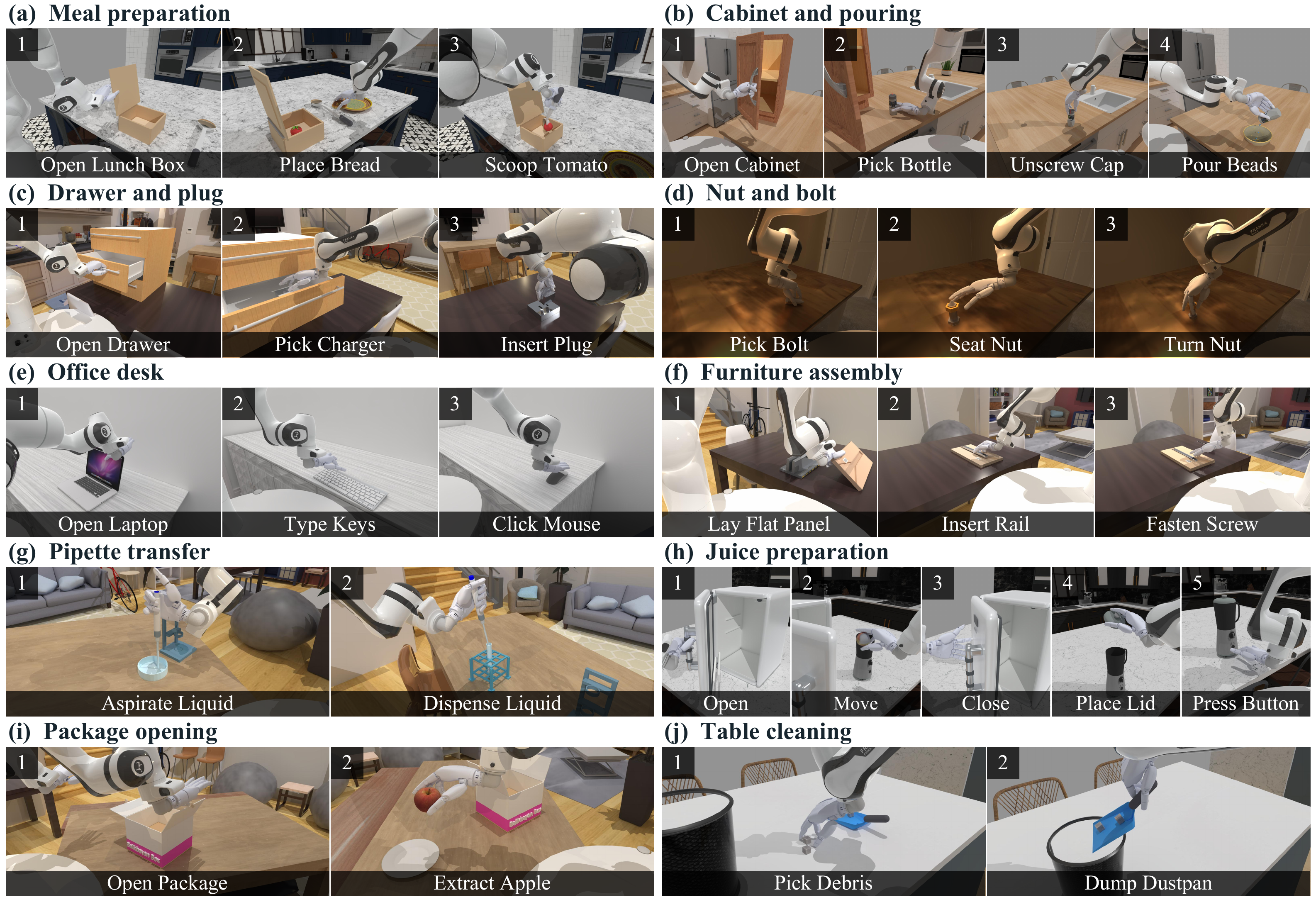}
    \caption{\textbf{Long-horizon tasks and their stages.}}
    \label{fig:10_tasks}
\end{figure}

Every task in DexArena is built on the same workstation, the same physics and the same parameter layers.
Only the objects, the layout and the success conditions are task-specific.
This matches each stage-level task with the corresponding long-horizon stage, and keeps the benchmark easy to extend.

\textbf{Tasks and their decomposition.}
Each long-horizon task is decomposed into two to five stages, as shown in \Cref{fig:10_tasks}.
The resulting 30 stages fall into six motion categories: \textit{Opening and Closing}, \textit{Grasping and Placement}, \textit{Insertion and Assembly}, \textit{Twisting and Screwing}, \textit{Pressing}, and \textit{Scooping and Pouring}.
The long-horizon task and its stages share the same layout, randomization ranges, and stage-level success conditions.
The workstation and the initial configuration of the arm are also shared.
For each stage-level task, only the arm(s) and hand(s) required by that task are retained in the scene.
All these settings allow a long-horizon policy to be evaluated stage by stage.
\begin{figure}[t]
    \centering
    \includegraphics[width=\linewidth]{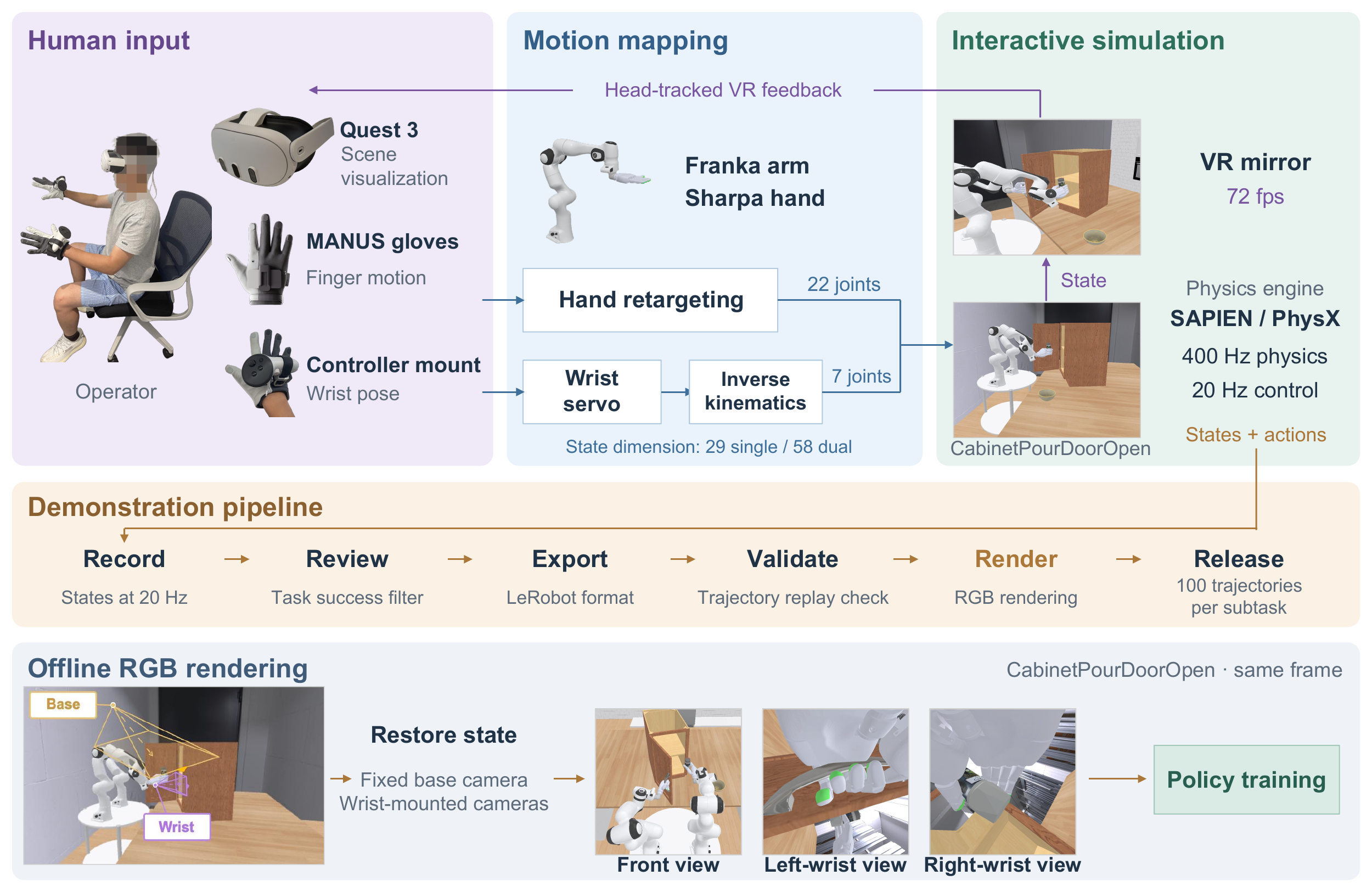}
    \caption{\textbf{Data collection pipeline.}
    The operator wears a VR headset and gloves, with a controller
    mounted on the back of each hand.
    Finger and wrist inputs are mapped to absolute joint targets.
    The scene is rendered in VR providing a better view for the operator.
    Each episode is recorded, then reviewed, exported, validated and rendered offline.}
    \label{fig:teleop_pipeline}
\end{figure}

\textbf{Success conditions.}
Every stage terminates when thresholds over physical quantities are satisfied, such as joint angles of articulated parts, relative poses, and insertion depths.
Conditions rely on live object poses, so they stay valid under randomization.
Most conditions must hold for 3 to 40 consecutive steps depending on the stage, so that an unstable contact is not mistaken for completion.

\textbf{Layered parameterization.}
Environment parameters are organized into four explicit layers.
The environment, hand and arm layers are fixed across all tasks.
Adding a task means updating only its task layer, without modifying the other three.
This allows a stage-level policy to be deployed unchanged inside the long-horizon task, and makes new tasks easy to build.

\textbf{Scenes, randomization and rendering.}
Initial object and robot poses are randomized at the beginning of each episode depending on the random seed.
Given the saved scene state of every frame, a trajectory can be re-rendered with different cameras while preserving the robot and object poses.
A stage-level task can also be evaluated inside a realistic indoor scene while preserving its layout and physics.

The details about task decomposition and parameterization are included in \Cref{subsec:task_info,subsec:parameters_environment}.

\subsection{Teleoperation Data Collection}
\label{subsec:teleoperation_data_collection}

\textbf{Teleoperation.}
The operator wears a Quest 3 headset and a pair of MANUS Meta Gloves Pro, with a Quest 3 controller mounted on the back of each hand for the wrist pose (Figure~\ref{fig:teleop_pipeline}). Finger motion is mapped to the 22 joints of the Sharpa hand by the official retargeting software \citep{sharpa2026manussdk}, which uses the same kinematic model as our simulated hand.
Other supported hands are retargeted with an open-source optimization-based library \citep{qin2023anyteleop}.
The wrist pose is converted to arm joint targets by inverse kinematics, with joint speeds capped at certain limits.
% The operator can freeze the fingers while repositioning the wrist, and sees markers showing where objects should go.
% The markers are visual only and are removed from every released camera view, so a policy cannot rely on them.
Recording runs at 20\,Hz and stops automatically one second after the task succeeds.
The action space is 29/58-dimensional for single/bimanual tasks, corresponding to absolute joint targets of the single arm/dual arms (7/14) and the single hand/dual hands (22/44).
Each frame stores the observation, the action, a sparse reward and episode flags in the LeRobot \citep{cadene2024lerobot} format; the scene state of every frame is kept alongside for restoration.
More details about teleoperation are in Appendix~\ref{subsec:teleop_system}.

\textbf{Validation and rendering.}
Episodes are reviewed after each session and only successful ones are kept.
Each dataset then goes through two checks.
First, recorded frames are restored in a new simulator instance and must reproduce the recorded scene state and rendered images, which makes offline rendering reliable. 
Camera views are rendered this way from the restored scene state: one fixed scene camera and one wrist camera per arm in use, each 256$\times$256.
Camera placement and rendering settings are given in Appendix~\ref{subsec:data_post-processing_validation_release}.
Second, the recorded actions are replayed open-loop from the recorded starting state.
In contact-rich phases such as grasping, insertion or pouring, tiny differences in the contact computation of the simulator can accumulate over time and may lead to failure during replay.
For each of the 30 stage-level tasks, we release 100 demonstrations that succeed on replay, yielding 3000 episodes and 9.1 hours in total, selected from 4112 accepted episodes.
More details about data post-processing are in Appendix~\ref{subsec:data_post-processing_validation_release}.

\subsection{Policy Learning}
\label{subsec:policy_learning}
Here, we briefly introduce how we adapt all the baseline methods for this benchmark.
Implementation details and training configurations for each method are included in \Cref{subsec:policy_implementations_training_configurations}.
The baseline methods can be mainly categorized into three types: Demonstration-guided Reinforcement Learning (RL), Imitation Learning (IL), and Vision-Language-Action (VLA) models.
All methods take image observations and robot proprioception as inputs.
For single-arm/bimanual tasks, visual observations consist of images from the front camera and the corresponding wrist camera(s).
Across all policy families, actions are represented as joint-space commands.

\paragraph{Demo-guided RL}
To avoid task-specific reward engineering, all DexArena tasks use sparse rewards defined directly by their success conditions.
However, training image-based RL policies from sparse rewards alone is challenging, particularly for dexterous manipulation.
We therefore incorporate human demonstrations into RL training.
For PPO \citep{schulman2017ppo}, we use three mechanisms to leverage the demonstrations.
First, we perform a Behavior Cloning (BC) warm-up and retain a BC loss term during subsequent RL training.
Second, we reset the environment from physical states sampled from demonstration trajectories with a certain probability, allowing the policy to progressively learn from states closer to successful behaviors.
Third, we initialize the action distribution of the policy and constrain its output range using the action statistics of the demonstrations.
We also evaluate Reinforcement Learning with Prior Data (RLPD) \citep{ball2023rlpd}, a demonstration-guided RL method that combines offline demonstrations with online experience throughout training.
Specifically, each training batch consists of an equal number of samples from the demonstration dataset and the online replay buffer.

\paragraph{Imitation Learning}
To adapt Action Chunking with Transformers (ACT) \citep{zhao2023act} to dexterous manipulation, we extend its state and action spaces to include all arm and hand joints.
To prevent the high-dimensional hand action from dominating optimization, we compute reconstruction losses separately for the arm and hand groups and assign a larger weight to the arm group.
We additionally condition the policy on multi-view temporal observations. 
We adapt Diffusion Policy (DP) \citep{chi2023diffusion} to the same dexterous-hand control interface by jointly generating absolute target positions for all arm and hand joints.
The policy is conditioned on the current robot joint positions and a short history of image observations.

\paragraph{VLA}
To complement the RL and imitation-learning baselines, we evaluate pretrained VLA models that additionally condition on language instructions.

We adapt the pretrained $\pi_{0.5}$ policy \citep{physicalintelligence2025pi05} to single- and bimanual dexterous manipulation by aligning its state and action interfaces with the 29- and 58-joint robot layouts and expanding the pretrained action projections where required.
To accommodate high-dimensional hand control, we predict absolute joint targets for both the arms and fingers and mask unavailable image views and unobserved or padded action channels.
Both standalone stage-level and long-horizon policies use LoRA~\citep{hu2022lora} fine-tuning, retaining the original flow-matching objective and predicting action chunks of length 30.

We also adapt GR00T \citep{gr00tn1_2025} to single- and bimanual dexterous manipulation by introducing new embodiment-specific state and action encoders, while retaining the original action-head dimensionality.
To account for the distinct characteristics of arm and hand control, we normalize their joint spaces separately and use a mixed action representation, with relative joint offsets for the arms and absolute joint targets for the fingers.
For GR00T, all model parameters, including the visual encoder and language backbone, are fine-tuned on the DexArena demonstrations.

For long-horizon evaluation, both VLA baselines use one task-specific model trained on pooled demonstrations from all constituent stages; stage transitions switch only the language instruction, while the same checkpoint remains fixed throughout each complete episode.
Implementation details and fine-tuning configurations are provided in Appendix~\ref{subsec:policy_implementations_training_configurations}.

\section{Experiments}
\label{sec:experiments}

As introduced in \Cref{subsec:policy_learning}, we study six policy learning baselines across individual-stage performance, generalization to varying scenarios, and long-horizon performance.
All models are trained using the teleoperation dataset from \Cref{subsec:teleoperation_data_collection}.
For completed stage-level and long-horizon measurements, we report the percentage of successful episodes from 50 evaluation episodes per task. Pending measurements are identified separately and do not constitute measured policy failures.
Task definitions, initial-state randomization and embodiments follow \Cref{subsec:task_design}. Method-specific training recipes, execution schedules and episode caps are documented in Appendix~\ref{subsec:policy_implementations_training_configurations}; these experiments are not compute-matched comparisons, and we do not assume a common episode cap where the recorded configurations differ.

\subsection{Teleoperation Dataset}
\label{subsec:teleoperation_dataset}
\begin{figure}[t]
    \centering
    \includegraphics[width=\linewidth]{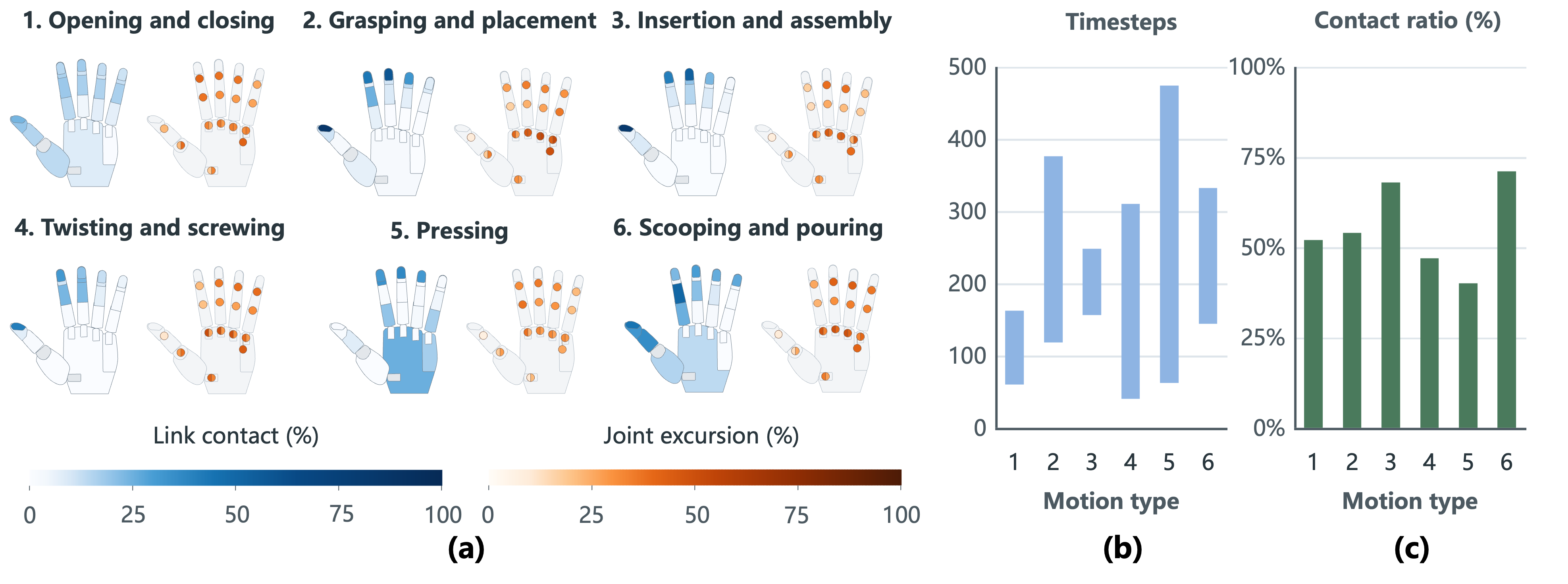}
    \caption{\textbf{Dataset Statistics Visualization}.
    Link contact: fraction of hand-contact frames.
    Joint excursion: motion amplitude of each hand joint averaged over episodes.
    Two degree-of-freedom joints are drawn as two half markers, the left half for bending and the right half for spreading.
    Timesteps: range of median episode lengths across tasks in each motion category.
    Contact ratio: fraction of frames with hand contact (any finger or the palm touching the primary object).}
    \label{fig:hand_usage}
\end{figure}

The released dataset contains 100 demonstrations for each of the 30 stage-level tasks, 3000 episodes and 9.1 hours in total. We analyze it from three angles:
hand usage, diversity of episodes and open-loop replay.
Figure~\ref{fig:hand_usage} shows that the six motion categories differ in hand-link contact, joint motion, episode length and contact ratio. 
Palm trajectories show substantial motion and inter-episode diversity.
The mean path length ranges from 0.35 to 2.04\,m, while the mean pairwise distance between palm trajectories ranges from 3.3 to 10.2\,cm across stage-level tasks.
Open-loop replay succeeds in 3222 of the 4112 accepted episodes, and tends to be less successful for more contact-rich tasks, from 98\% for opening and closing to 65\% for insertion and assembly.
Definitions and per stage-level task results are given in \Cref{subsec:dataset_statistics_and_analysis}.

\subsection{Stage-level Task Performance}
\label{subsec:stage-level-tasks-performance}

We first train and evaluate the stage-level baselines on all 30 tasks individually.
Each episode starts from a valid initial state for the corresponding stage, and success is determined by its physically grounded completion criterion.
Table~\ref{tab:stage-level-tasks-performance} reports the success rate for every policy on each stage.

\begin{table*}[!t]
\centering
\caption{\textbf{Stage-level task performance.} Success rates (\%, $\uparrow$) on all 30 stage-level tasks in the training scene shown in \Cref{fig:generalization_evaluation_scenes}, grouped by their long-horizon task, with 50 evaluation episodes conducted per task.}
\label{tab:stage-level-tasks-performance}
\fontsize{8.2}{9.4}\selectfont
\setlength{\tabcolsep}{4pt}
\renewcommand{\arraystretch}{1.10}

\begin{tabularx}{\textwidth}{@{}>{\raggedright\arraybackslash}p{0.20\textwidth}|>{\raggedright\arraybackslash}p{0.16\textwidth}|*{2}{>{\centering\arraybackslash}X}|*{2}{>{\centering\arraybackslash}X}|*{2}{>{\centering\arraybackslash}X}@{}}
\toprule
\multirow[c]{2}{=}{\raggedright\textbf{Long-horizon}}
& \multirow[c]{2}{=}{\raggedright\textbf{Stages}}
& \multicolumn{2}{c|}{\textbf{Demo-guided RL}}
& \multicolumn{2}{c|}{\textbf{Imitation learning}}
& \multicolumn{2}{c}{\textbf{VLA models}} \\
\cmidrule(lr){3-4}\cmidrule(lr){5-6}\cmidrule(l){7-8}
& & \textbf{PPO} & \textbf{RLPD}
& \textbf{ACT} & \textbf{DP}
& \textbf{GR00T} & \shortstack{$\boldsymbol{\pi_{0.5}}$\\\textbf{(LoRA)}} \\
\midrule
\multirow{3}{*}{\textbf{Meal preparation}}
& \cellcolor{tablehead}OpenLunchBox & \cellcolor{tablehead}74 & \cellcolor{tablehead}100 & \cellcolor{tablehead}100 & \cellcolor{tablehead}96 & \cellcolor{tablehead}100 & \cellcolor{tablehead}100 \\
& \cellcolor{tablehead}PlaceBread & \cellcolor{tablehead}12 & \cellcolor{tablehead}86 & \cellcolor{tablehead}88 & \cellcolor{tablehead}20 & \cellcolor{tablehead}60 & \cellcolor{tablehead}70 \\
& \cellcolor{tablehead}ScoopTomato & \cellcolor{tablehead}0 & \cellcolor{tablehead}0 & \cellcolor{tablehead}44 & \cellcolor{tablehead}0 & \cellcolor{tablehead}12 & \cellcolor{tablehead}32 \\
\addlinespace[2pt]
\multirow{4}{*}{\textbf{Cabinet and pouring}}
& OpenCabinet & 62 & 100 & 94 & 88 & 92 & 98 \\
& PickBottle & 0 & 14 & 68 & 18 & 78 & 54 \\
& UnscrewCap & 0 & 2 & 62 & 50 & 62 & 46 \\
& PourBeads & 0 & 96 & 100 & 32 & 78 & 52 \\
\addlinespace[2pt]
\multirow{3}{*}{\textbf{Drawer and plug}}
& \cellcolor{tablehead}OpenDrawer & \cellcolor{tablehead}98 & \cellcolor{tablehead}100 & \cellcolor{tablehead}98 & \cellcolor{tablehead}98 & \cellcolor{tablehead}98 & \cellcolor{tablehead}98 \\
& \cellcolor{tablehead}PickCharger & \cellcolor{tablehead}0 & \cellcolor{tablehead}62 & \cellcolor{tablehead}92 & \cellcolor{tablehead}0 & \cellcolor{tablehead}78 & \cellcolor{tablehead}78 \\
& \cellcolor{tablehead}InsertPlug & \cellcolor{tablehead}0 & \cellcolor{tablehead}0 & \cellcolor{tablehead}16 & \cellcolor{tablehead}12 & \cellcolor{tablehead}6 & \cellcolor{tablehead}12 \\
\addlinespace[2pt]
\multirow{3}{*}{\textbf{Nut and bolt}}
& PickBolt & 0 & 28 & 8 & 22 & 44 & 12 \\
& SeatNut & 0 & 34 & 12 & 6 & 26 & 2 \\
& TurnNut & 76 & 100 & 96 & 100 & 100 & 100 \\
\addlinespace[2pt]
\multirow{3}{*}{\textbf{Office desk}}
& \cellcolor{tablehead}OpenLaptop & \cellcolor{tablehead}100 & \cellcolor{tablehead}100 & \cellcolor{tablehead}100 & \cellcolor{tablehead}88 & \cellcolor{tablehead}100 & \cellcolor{tablehead}82 \\
& \cellcolor{tablehead}TypeKeys & \cellcolor{tablehead}0 & \cellcolor{tablehead}0 & \cellcolor{tablehead}0 & \cellcolor{tablehead}0 & \cellcolor{tablehead}32 & \cellcolor{tablehead}0 \\
& \cellcolor{tablehead}ClickMouse & \cellcolor{tablehead}96 & \cellcolor{tablehead}100 & \cellcolor{tablehead}14 & \cellcolor{tablehead}92 & \cellcolor{tablehead}98 & \cellcolor{tablehead}62 \\
\addlinespace[2pt]
\multirow{3}{*}{\textbf{Furniture assembly}}
& LayFlatPanel & 0 & 96 & 78 & 24 & 64 & 0 \\
& InsertRail & 0 & 0 & 22 & 2 & 16 & 8 \\
& FastenScrew & 100 & 100 & 0 & 0 & 84 & 94 \\
\addlinespace[2pt]
\multirow{2}{*}{\textbf{Pipette transfer}}
& \cellcolor{tablehead}AspirateLiquid & \cellcolor{tablehead}0 & \cellcolor{tablehead}0 & \cellcolor{tablehead}40 & \cellcolor{tablehead}2 & \cellcolor{tablehead}8 & \cellcolor{tablehead}26 \\
& \cellcolor{tablehead}DispenseLiquid & \cellcolor{tablehead}0 & \cellcolor{tablehead}0 & \cellcolor{tablehead}0 & \cellcolor{tablehead}2 & \cellcolor{tablehead}8 & \cellcolor{tablehead}0 \\
\addlinespace[2pt]
\multirow{5}{*}{\textbf{Juice preparation}}
& OpenFridge & 90 & 98 & 100 & 88 & 94 & 96 \\
& AppleToBlender & 0 & 0 & 92 & 42 & 54 & 38 \\
& CloseFridge & 100 & 100 & 100 & 100 & 98 & 100 \\
& PlaceLid & 0 & 78 & 26 & 26 & 42 & 16 \\
& PressButton & 26 & 100 & 66 & 72 & 64 & 68 \\
\addlinespace[2pt]
\multirow{2}{*}{\textbf{Package opening}}
& \cellcolor{tablehead}OpenPackage & \cellcolor{tablehead}2 & \cellcolor{tablehead}92 & \cellcolor{tablehead}44 & \cellcolor{tablehead}2 & \cellcolor{tablehead}64 & \cellcolor{tablehead}58 \\
& \cellcolor{tablehead}ExtractApple & \cellcolor{tablehead}0 & \cellcolor{tablehead}42 & \cellcolor{tablehead}88 & \cellcolor{tablehead}22 & \cellcolor{tablehead}100 & \cellcolor{tablehead}82 \\
\addlinespace[2pt]
\multirow{2}{*}{\textbf{Table cleaning}}
& PickDebris & 0 & 0 & 0 & 2 & 0 & 0 \\
& DumpDustpan & 0 & 0 & 92 & 70 & 82 & 78 \\
\midrule
\rowcolor{tablesummary}
\multicolumn{2}{l}{\textbf{Average}} & 27.9 & 54.3 & 58.0 & 39.2 & 61.4 & 52.1 \\
\bottomrule
\end{tabularx}
\end{table*}

\subsection{Generalization Performance in Varying Visual Scenes}
\label{subsec:generalization-performance}

We next evaluate the available trained stage-level policies on the same 30 tasks under three different scenarios, as shown in \Cref{appendix:generalization_scenes}, without additional training or adaptation.
For each stage and policy, Table~\ref{tab:generalization-performance} reports the average success rate across the evaluation scenarios.
The final row shows average scores over all 30 stage-level tasks.
Comparison with Table~\ref{tab:stage-level-tasks-performance} quantifies the change in performance under scenario variation.

\begin{table*}[!t]
\centering
\caption{\textbf{Generalization performance.} Success rates (\%, $\uparrow$) on the same 30 stage-level tasks under varying scenarios, grouped by their long-horizon task. Results are reported as mean$_{\pm\mathrm{std}}$ across the three held-out scenes, with 50 evaluation episodes conducted per scene.}
\label{tab:generalization-performance}
\fontsize{8.2}{9.4}\selectfont
\setlength{\tabcolsep}{4pt}
\renewcommand{\arraystretch}{1.10}

\begin{tabularx}{\textwidth}{@{}>{\raggedright\arraybackslash}p{0.18\textwidth}|>{\raggedright\arraybackslash}p{0.13\textwidth}|*{2}{>{\centering\arraybackslash}X}|*{2}{>{\centering\arraybackslash}X}|*{2}{>{\centering\arraybackslash}X}@{}}
\toprule
\multirow[c]{2}{*}{\textbf{Long-horizon}} & \multirow[c]{2}{*}{\textbf{Stages}} & \multicolumn{2}{c|}{\textbf{Demo-guided RL}} & \multicolumn{2}{c|}{\textbf{Imitation learning}} & \multicolumn{2}{c}{\textbf{VLA models}} \\
\cmidrule(lr){3-4}\cmidrule(lr){5-6}\cmidrule(l){7-8}
& & \textbf{PPO} & \textbf{RLPD} & \textbf{ACT} & \textbf{DP} & \textbf{GR00T} & \shortstack{$\boldsymbol{\pi_{0.5}}$\\\textbf{(LoRA)}} \\
\midrule
\multirow{3}{*}{\textbf{Meal preparation}}
& \cellcolor{tablehead}OpenLunchBox & \cellcolor{tablehead}$35.3_{\pm 14.5}$ & \cellcolor{tablehead}$33.3_{\pm 57.7}$ & \cellcolor{tablehead}$100.0_{\pm 0.0}$ & \cellcolor{tablehead}$35.3_{\pm 31.6}$ & \cellcolor{tablehead}$100.0_{\pm 0.0}$ & \cellcolor{tablehead}$99.3_{\pm 1.2}$ \\
& \cellcolor{tablehead}PlaceBread & \cellcolor{tablehead}$10.0_{\pm 3.5}$ & \cellcolor{tablehead}$0.0_{\pm 0.0}$ & \cellcolor{tablehead}$40.0_{\pm 30.0}$ & \cellcolor{tablehead}$8.7_{\pm 4.6}$ & \cellcolor{tablehead}$24.0_{\pm 26.2}$ & \cellcolor{tablehead}$71.3_{\pm 7.0}$ \\
& \cellcolor{tablehead}ScoopTomato & \cellcolor{tablehead}$0.0_{\pm 0.0}$ & \cellcolor{tablehead}$0.0_{\pm 0.0}$ & \cellcolor{tablehead}$23.3_{\pm 9.0}$ & \cellcolor{tablehead}$0.0_{\pm 0.0}$ & \cellcolor{tablehead}$6.7_{\pm 6.1}$ & \cellcolor{tablehead}$18.7_{\pm 3.1}$ \\
\addlinespace[2pt]
\multirow{4}{*}{\textbf{Cabinet and pouring}}
& OpenCabinet & $8.7_{\pm 9.0}$ & $63.3_{\pm 54.9}$ & $93.3_{\pm 6.4}$ & $84.0_{\pm 11.1}$ & $78.7_{\pm 17.9}$ & $97.3_{\pm 2.3}$ \\
& PickBottle & $0.0_{\pm 0.0}$ & $0.0_{\pm 0.0}$ & $76.0_{\pm 8.0}$ & $4.0_{\pm 5.3}$ & $46.0_{\pm 36.7}$ & $69.3_{\pm 6.4}$ \\
& UnscrewCap & $0.0_{\pm 0.0}$ & $0.0_{\pm 0.0}$ & $4.0_{\pm 3.5}$ & $1.3_{\pm 2.3}$ & $9.3_{\pm 14.5}$ & $33.3_{\pm 11.0}$ \\
& PourBeads & $0.7_{\pm 1.2}$ & $0.0_{\pm 0.0}$ & $42.0_{\pm 22.3}$ & $6.0_{\pm 10.4}$ & $16.7_{\pm 17.5}$ & $36.0_{\pm 5.3}$ \\
\addlinespace[2pt]
\multirow{3}{*}{\textbf{Drawer and plug}}
& \cellcolor{tablehead}OpenDrawer & \cellcolor{tablehead}$97.3_{\pm 4.6}$ & \cellcolor{tablehead}$95.3_{\pm 4.2}$ & \cellcolor{tablehead}$82.7_{\pm 10.1}$ & \cellcolor{tablehead}$4.0_{\pm 5.3}$ & \cellcolor{tablehead}$90.0_{\pm 10.6}$ & \cellcolor{tablehead}$100.0_{\pm 0.0}$ \\
& \cellcolor{tablehead}PickCharger & \cellcolor{tablehead}$0.0_{\pm 0.0}$ & \cellcolor{tablehead}$5.3_{\pm 9.2}$ & \cellcolor{tablehead}$88.0_{\pm 5.3}$ & \cellcolor{tablehead}$0.0_{\pm 0.0}$ & \cellcolor{tablehead}$64.7_{\pm 19.6}$ & \cellcolor{tablehead}$88.0_{\pm 0.0}$ \\
& \cellcolor{tablehead}InsertPlug & \cellcolor{tablehead}$0.0_{\pm 0.0}$ & \cellcolor{tablehead}$0.0_{\pm 0.0}$ & \cellcolor{tablehead}$0.7_{\pm 1.2}$ & \cellcolor{tablehead}$0.0_{\pm 0.0}$ & \cellcolor{tablehead}$2.0_{\pm 3.5}$ & \cellcolor{tablehead}$1.3_{\pm 1.2}$ \\
\addlinespace[2pt]
\multirow{3}{*}{\textbf{Nut and bolt}}
& PickBolt & $0.7_{\pm 1.2}$ & $0.0_{\pm 0.0}$ & $8.7_{\pm 5.0}$ & $2.0_{\pm 3.5}$ & $21.3_{\pm 13.3}$ & $18.0_{\pm 2.0}$ \\
& SeatNut & $0.0_{\pm 0.0}$ & $0.0_{\pm 0.0}$ & $2.7_{\pm 2.3}$ & $0.7_{\pm 1.2}$ & $2.0_{\pm 3.5}$ & $3.3_{\pm 2.3}$ \\
& TurnNut & $26.7_{\pm 11.0}$ & $11.3_{\pm 16.3}$ & $44.0_{\pm 19.7}$ & $11.3_{\pm 12.7}$ & $80.0_{\pm 19.7}$ & $87.3_{\pm 5.0}$ \\
\addlinespace[2pt]
\multirow{3}{*}{\textbf{Office desk}}
& \cellcolor{tablehead}OpenLaptop & \cellcolor{tablehead}$99.3_{\pm 1.2}$ & \cellcolor{tablehead}$48.0_{\pm 50.1}$ & \cellcolor{tablehead}$84.7_{\pm 18.1}$ & \cellcolor{tablehead}$88.7_{\pm 3.1}$ & \cellcolor{tablehead}$95.3_{\pm 8.1}$ & \cellcolor{tablehead}$66.7_{\pm 5.0}$ \\
& \cellcolor{tablehead}TypeKeys & \cellcolor{tablehead}$0.0_{\pm 0.0}$ & \cellcolor{tablehead}$0.0_{\pm 0.0}$ & \cellcolor{tablehead}$0.0_{\pm 0.0}$ & \cellcolor{tablehead}$0.0_{\pm 0.0}$ & \cellcolor{tablehead}$17.3_{\pm 17.0}$ & \cellcolor{tablehead}$0.0_{\pm 0.0}$ \\
& \cellcolor{tablehead}ClickMouse & \cellcolor{tablehead}$98.0_{\pm 3.5}$ & \cellcolor{tablehead}$6.7_{\pm 11.5}$ & \cellcolor{tablehead}$12.7_{\pm 4.6}$ & \cellcolor{tablehead}$68.7_{\pm 15.0}$ & \cellcolor{tablehead}$100.0_{\pm 0.0}$ & \cellcolor{tablehead}$64.7_{\pm 3.1}$ \\
\addlinespace[2pt]
\multirow{3}{*}{\textbf{Furniture assembly}}
& LayFlatPanel & $0.0_{\pm 0.0}$ & $0.0_{\pm 0.0}$ & $54.0_{\pm 27.5}$ & $0.0_{\pm 0.0}$ & $34.7_{\pm 28.4}$ & $0.0_{\pm 0.0}$ \\
& InsertRail & $0.0_{\pm 0.0}$ & $0.0_{\pm 0.0}$ & $0.0_{\pm 0.0}$ & $0.0_{\pm 0.0}$ & $6.0_{\pm 10.4}$ & $9.3_{\pm 3.1}$ \\
& FastenScrew & $86.0_{\pm 24.2}$ & $70.7_{\pm 50.8}$ & $85.3_{\pm 11.5}$ & $0.0_{\pm 0.0}$ & $20.0_{\pm 26.2}$ & $94.0_{\pm 3.5}$ \\
\addlinespace[2pt]
\multirow{2}{*}{\textbf{Pipette transfer}}
& \cellcolor{tablehead}AspirateLiquid & \cellcolor{tablehead}$0.0_{\pm 0.0}$ & \cellcolor{tablehead}$0.0_{\pm 0.0}$ & \cellcolor{tablehead}$8.7_{\pm 8.1}$ & \cellcolor{tablehead}$1.3_{\pm 1.2}$ & \cellcolor{tablehead}$2.0_{\pm 2.0}$ & \cellcolor{tablehead}$34.0_{\pm 6.9}$ \\
& \cellcolor{tablehead}DispenseLiquid & \cellcolor{tablehead}$0.0_{\pm 0.0}$ & \cellcolor{tablehead}$0.0_{\pm 0.0}$ & \cellcolor{tablehead}$0.0_{\pm 0.0}$ & \cellcolor{tablehead}$0.0_{\pm 0.0}$ & \cellcolor{tablehead}$2.0_{\pm 2.0}$ & \cellcolor{tablehead}$4.0_{\pm 0.0}$ \\
\addlinespace[2pt]
\multirow{5}{*}{\textbf{Juice preparation}}
& OpenFridge & $65.3_{\pm 22.0}$ & $78.7_{\pm 35.2}$ & $92.7_{\pm 9.5}$ & $68.0_{\pm 33.3}$ & $70.0_{\pm 45.1}$ & $96.0_{\pm 3.5}$ \\
& AppleToBlender & $0.7_{\pm 1.2}$ & $0.0_{\pm 0.0}$ & $71.3_{\pm 14.2}$ & $7.3_{\pm 12.7}$ & $18.0_{\pm 2.0}$ & $38.7_{\pm 4.2}$ \\
& CloseFridge & $100.0_{\pm 0.0}$ & $100.0_{\pm 0.0}$ & $99.3_{\pm 1.2}$ & $66.0_{\pm 47.3}$ & $92.0_{\pm 8.7}$ & $100.0_{\pm 0.0}$ \\
& PlaceLid & $0.0_{\pm 0.0}$ & $0.0_{\pm 0.0}$ & $18.0_{\pm 11.1}$ & $1.3_{\pm 2.3}$ & $28.7_{\pm 12.9}$ & $16.0_{\pm 2.0}$ \\
& PressButton & $18.0_{\pm 5.3}$ & $18.7_{\pm 32.3}$ & $54.0_{\pm 19.3}$ & $52.7_{\pm 40.6}$ & $42.7_{\pm 10.1}$ & $62.0_{\pm 7.2}$ \\
\addlinespace[2pt]
\multirow{2}{*}{\textbf{Package opening}}
& \cellcolor{tablehead}OpenPackage & \cellcolor{tablehead}$2.0_{\pm 0.0}$ & \cellcolor{tablehead}$2.7_{\pm 4.6}$ & \cellcolor{tablehead}$36.7_{\pm 22.1}$ & \cellcolor{tablehead}$1.3_{\pm 2.3}$ & \cellcolor{tablehead}$35.3_{\pm 22.1}$ & \cellcolor{tablehead}$30.7_{\pm 24.8}$ \\
& \cellcolor{tablehead}ExtractApple & \cellcolor{tablehead}$0.0_{\pm 0.0}$ & \cellcolor{tablehead}$0.0_{\pm 0.0}$ & \cellcolor{tablehead}$78.7_{\pm 3.1}$ & \cellcolor{tablehead}$3.3_{\pm 3.1}$ & \cellcolor{tablehead}$54.7_{\pm 39.5}$ & \cellcolor{tablehead}$70.0_{\pm 26.5}$ \\
\addlinespace[2pt]
\multirow{2}{*}{\textbf{Table cleaning}}
& PickDebris & $1.3_{\pm 2.3}$ & $0.0_{\pm 0.0}$ & $0.0_{\pm 0.0}$ & $0.0_{\pm 0.0}$ & $1.3_{\pm 1.2}$ & $0.0_{\pm 0.0}$ \\
& DumpDustpan & $0.7_{\pm 1.2}$ & $0.0_{\pm 0.0}$ & $59.3_{\pm 17.9}$ & $1.3_{\pm 2.3}$ & $88.7_{\pm 2.3}$ & $68.0_{\pm 4.0}$ \\
\midrule
\rowcolor{tablesummary}
\multicolumn{2}{l}{\textbf{Average}}
& $\mathbf{21.7_{\pm 3.5}}$
& $\mathbf{17.8_{\pm 10.9}}$
& $\mathbf{45.4_{\pm 3.6}}$
& $\mathbf{17.2_{\pm 5.7}}$
& $\mathbf{41.7_{\pm 14.2}}$
& $\mathbf{49.2_{\pm 2.3}}$ \\
\bottomrule
\end{tabularx}
\end{table*}

\subsection{Long-Horizon Performance}
\label{subsec:long-horizon-performance}
The long-horizon protocol covers six baselines on ten tasks. For PPO, RLPD, ACT and DP, we sequentially execute the learned stage-level policies and switch policies when the current stage's success condition is met. For both VLA baselines ($\pi_{0.5}$ and GR00T), one task-specific model is trained on pooled demonstrations from all constituent stages. One fixed checkpoint controls the full episode; stage transitions switch only the language instruction, not model weights. The environment is initialized only at the beginning and is not reset between stages. Success requires every stage to complete in order and the native full-task success condition to hold within the evaluation horizon.

Among completed evaluations, long-horizon success remains low across most tasks, indicating that individual-stage competence does not reliably translate into successful full-task execution. Per-task results and evaluation status are provided in Appendix~\ref{appendix:long_horizon_performance}.

% \begin{table*}[!t]
% \centering
% \caption{\textbf{Average long-horizon performance.} Average end-to-end success rate (\%, $\uparrow$) across 10 tasks using sequentially composed stage-level policies.}
% \label{tab:long-horizon-average}
% \fontsize{8.2}{9.4}\selectfont
% \setlength{\tabcolsep}{10pt}
% \renewcommand{\arraystretch}{1.25}

% \begin{tabular*}{\textwidth}{@{\extracolsep{\fill}}lccccc@{}}
% \toprule
% \textbf{Metric} & \textbf{PPO} & \textbf{RLPD} & \textbf{ACT} & \textbf{DP} & \textbf{GR00T} \\
% \midrule
% \rowcolor{tablesummary}
% \textbf{Average} & 0 & 0 & 1.2 & 0 & 0.6 \\
% \bottomrule
% \end{tabular*}

% \end{table*}

\section{Conclusion}

We introduced DexArena, a benchmark for studying long-horizon dexterous manipulation through 10 multi-stage tasks and 30 physically grounded stage-level tasks.
The shared task and stage definitions, four end-effectors, replayable simulation, and teleoperated demonstrations support evaluation of individual skills, scene variation, and sequential task completion in a common framework.
The baseline results show that performance on isolated stage-level tasks does not by itself yield reliable long-horizon behavior.
Performance also varies substantially across stages and evaluation scenes, exposing persistent challenges in contact-rich manipulation and robustness to visual variation.

\section{Limitation}
Although we calibrate the simulator parameters to better match real-world robot dynamics, accurately reproducing the complex contact dynamics of dexterous manipulation remains challenging.
In particular, contacts between multiple fingers, objects, and articulated components can be sensitive to small discrepancies in geometry, friction, and actuation, leaving a non-negligible sim-to-real gap.
Further reducing this gap will require more extensive system identification and validation against physical-robot interactions.

\clearpage

\ificlrversion
\subsection*{AI use statement}
We used generative AI tools to assist with implementing the simulation tasks, reviewing experimental methodology and results, and polishing the language of the manuscript.
All AI-assisted code, analyses, and text were reviewed, and the simulation implementation was tested by the authors.
We take responsibility for the final content of this work, including all text, claims, code, data, and other artifacts.

\subsection*{Ethics statement}
We have considered the issues covered by the ICLR Code of Ethics and do not
identify any ethical concerns raised by this work.

\subsection*{Reproducibility statement}
We will release the complete source code, datasets, training and evaluation scripts, and trained model checkpoints.
The main paper and appendix document the task construction, teleoperation data collection and
validation, policy implementations, training configurations, and evaluation
protocols (see Sections~\ref{subsec:teleoperation_data_collection} and
\ref{sec:experiments}, and Appendices~\ref{subsec:data_post-processing_validation_release}
and~\ref{subsec:policy_implementations_training_configurations}).
% Moreover, we configure the random seeds explicitly for both training and evaluation, report the seeds used for the experiments, and include them in the released configurations so that the reported results can be reproduced.
Moreover, we configure the random seeds explicitly for both training and evaluation and include them in the released configurations so that the reported results can be reproduced.
\fi



\ificlrversion
    \bibliographystyle{iclr2027_conference}
\else
    \bibliographystyle{unsrtnat}
\fi
\bibliography{iclr2027_conference}

\clearpage

\appendix
\section{Appendix}
\subsection{Related Work on Policy Learning}
\label{subsec:related_work_policy_learning}

Data-driven policy learning has become a dominant paradigm for robot manipulation, with imitation learning (IL) and reinforcement learning (RL) providing complementary approaches to acquiring manipulation skills.
Early data-driven methods primarily focused on learning single-task policies from demonstrations.
Action Chunking with Transformers (ACT) ~\citep{zhao2023act} predicts temporally coherent action sequences from demonstrations, while Diffusion Policy (DP) ~\citep{chi2023diffusion} models multi-modal action distributions through iterative denoising.
These methods have demonstrated strong performance in individual manipulation tasks, but their training and generalization are typically constrained by the scale and diversity of the collected demonstrations.

Recent research has increasingly moved toward multi-task and large-scale pretrained policies, enabled by the availability of diverse robot datasets and large pretrained models.
Vision-Language-Action (VLA) models incorporate visual and language knowledge to learn general-purpose manipulation behaviors across tasks and embodiments.
For example, $\pi_0$ ~\citep{black2024pi0} builds a flow-based action expert based pre-trained Vision-Language Model (VLM) for general robot control, while GR00T ~\citep{gr00tn1_2025} develops a hierarchical foundation model for generalist robot manipulation.
Beyond VLA policies, emerging approaches such as World-Action-Model ~\citep{ye2026worldactionmodelszeroshot} further incorporate video-generative models to learn predictive representations of the environment and support video-pre-training and policy learning in a unified model architecture.
Building on $\pi_0$, $\pi_{0.5}$ \citep{physicalintelligence2025pi05} co-trains on heterogeneous robot and vision-language data, combining high-level semantic supervision with low-level action learning to improve open-world generalization. We use its pretrained policy as an initialization for dexterous manipulation. LoRA~\citep{hu2022lora} introduces trainable low-rank weight updates; our adaptation combines these updates with trainable visual and action-interface modules, rather than freezing the entire remaining network. Our DexArena LoRA adaptation applies to both standalone stage-level and long-horizon policies; the latter additionally pools constituent-stage demonstrations. The adaptation and training protocols are described in Section~\ref{subsec:policy_learning} and Appendix~\ref{app:pi05_pooled_protocol}.

In parallel, reinforcement learning provides an important mechanism for learning and adapting policies through environment interaction, particularly in simulation where large amounts of interaction data can be collected efficiently. PPO employs clipped on-policy optimization for policy improvement~\citep{schulman2017ppo}, while RLPD combines demonstrations with online experience through replay-based learning~\citep{ball2023rlpd}.
RL can therefore serve both as a standalone policy-learning paradigm and as a complementary approach for post-training pretrained or imitation-learned policies, allowing them to further optimize task-specific objectives through interaction.

These developments motivate evaluating manipulation policies across different levels of data scale, prior knowledge, and interaction. 
DexArena provides a common evaluation framework for comparing such data-driven approaches under long-horizon dexterous manipulation, where performance depends not only on learning individual skills but also on composing them over multiple stages and handling diverse embodiments and interaction dynamics.

\subsection{Task Information}
\label{subsec:task_info}

Table~\ref{tab:taxonomy} groups the 30 stage-level tasks by the dominant motion they require.

\begin{table}[t]
\centering\small
\caption{The 30 stage-level tasks by motion category, with properties of the released
demonstrations (100 per stage-level task). Time to success: range of the stage-level task medians.
Contact is estimated by re-simulating each recorded frame (threshold 0.1\,N).
Contact ratio: fraction of frames, up to first success, in which a finger or
the palm touches the primary object of the stage-level task. Fingers: number of distinct fingers in contact, averaged over hand-contact frames. Stage-level task values pool the frames of its 100 episodes; category values weight stage-level tasks equally.}
\label{tab:taxonomy}
\setlength{\tabcolsep}{2.5pt}
\begin{tabular}{@{}lc@{\hspace{0.8em}}ll@{\hspace{0.8em}}ccc@{}}
\toprule
Category & \# & \multicolumn{2}{l}{Stages} & Time (s) & Contact ratio & Fingers \\
\midrule
\multirow{4}{*}{Opening and closing} & \multirow{4}{*}{7}
  & \texttt{\footnotesize OpenCabinet}    & \texttt{\footnotesize OpenDrawer}     & \multirow{4}{*}{3.0--8.1}  & \multirow{4}{*}{0.52} & \multirow{4}{*}{2.3} \\
& & \texttt{\footnotesize OpenFridge}     & \texttt{\footnotesize CloseFridge}    & & & \\
& & \texttt{\footnotesize OpenLunchBox}   & \texttt{\footnotesize OpenLaptop}     & & & \\
& & \texttt{\footnotesize OpenPackage}    &                                       & & & \\
\addlinespace
\multirow{4}{*}{Grasping and placement} & \multirow{4}{*}{8}
  & \texttt{\footnotesize PickBottle}     & \texttt{\footnotesize PickCharger}    & \multirow{4}{*}{5.9--18.8} & \multirow{4}{*}{0.54} & \multirow{4}{*}{3.1} \\
& & \texttt{\footnotesize PickBolt}       & \texttt{\footnotesize PlaceBread}     & & & \\
& & \texttt{\footnotesize ExtractApple}   & \texttt{\footnotesize AppleToBlender} & & & \\
& & \texttt{\footnotesize LayFlatPanel}   & \texttt{\footnotesize PickDebris}     & & & \\
\addlinespace
\multirow{2}{*}{Insertion and assembly} & \multirow{2}{*}{4}
  & \texttt{\footnotesize InsertPlug}     & \texttt{\footnotesize InsertRail}     & \multirow{2}{*}{7.8--12.4} & \multirow{2}{*}{0.68} & \multirow{2}{*}{2.7} \\
& & \texttt{\footnotesize SeatNut}        & \texttt{\footnotesize PlaceLid}       & & & \\
\addlinespace
\multirow{2}{*}{Twisting and screwing} & \multirow{2}{*}{3}
  & \texttt{\footnotesize UnscrewCap}     & \texttt{\footnotesize TurnNut}        & \multirow{2}{*}{2.0--15.5} & \multirow{2}{*}{0.47} & \multirow{2}{*}{2.1} \\
& & \texttt{\footnotesize FastenScrew}    &                                       & & & \\
\addlinespace
\multirow{3}{*}{Pressing} & \multirow{3}{*}{5}
  & \texttt{\footnotesize PressButton}    & \texttt{\footnotesize ClickMouse}     & \multirow{3}{*}{3.1--23.7} & \multirow{3}{*}{0.40} & \multirow{3}{*}{1.9} \\
& & \texttt{\footnotesize TypeKeys}       & \texttt{\footnotesize AspirateLiquid} & & & \\
& & \texttt{\footnotesize DispenseLiquid} &                                       & & & \\
\addlinespace
\multirow{2}{*}{Scooping and pouring} & \multirow{2}{*}{3}
  & \texttt{\footnotesize ScoopTomato}    & \texttt{\footnotesize PourBeads}      & \multirow{2}{*}{7.2--16.6} & \multirow{2}{*}{0.71} & \multirow{2}{*}{3.1} \\
& & \texttt{\footnotesize DumpDustpan}    &                                       & & & \\
\bottomrule
\end{tabular}
\end{table}

\Cref{tab:subtask_specs} lists every stage-level task with the hand it uses, its episode horizon, and the condition under which it is credited as complete.

\footnotesize
\setlength{\tabcolsep}{3pt}
\begin{xltabular}{\linewidth}{@{}l c r >{\raggedright\arraybackslash}X >{\raggedright\arraybackslash}X@{}}
\caption{The 30 stage-level tasks: hand used, episode horizon in control steps, description,
and success condition. ``Hold $n$'' means the condition must stay true for
$n$ consecutive control steps.}
\label{tab:subtask_specs} \\
\toprule
Stages & Hand & Horizon & Description & Success condition \\
\midrule
\endfirsthead
\multicolumn{5}{@{}l}{\textit{\Cref{tab:subtask_specs}, continued}} \\
\toprule
Stages & Hand & Horizon & Description & Success condition \\
\midrule
\endhead
\midrule
\multicolumn{5}{r@{}}{\textit{continued on next page}} \\
\endfoot
\bottomrule
\endlastfoot
\multicolumn{5}{@{}l}{Opening and closing} \\
\midrule
OpenLunchBox & L & 300 & Open the lid of a lunch box. & Lid angle $\geq 90^\circ$, opened by the left hand; hold 20. \\
OpenCabinet & L & 600 & Open a cabinet door. & Door angle $\geq 80^\circ$; hold 5. \\
OpenDrawer & L & 600 & Pull a drawer open. & Drawer travel $> 0.18$\,m; hold 5. \\
OpenFridge & L & 650 & Open a mini-fridge door. & Door angle $\geq 75^\circ$ and door speed $\leq 0.12$\,rad/s; hold 6. \\
CloseFridge & L & 650 & Close the fridge door from $75^\circ$. & Door angle $\leq 1^\circ$ and speed $\leq 0.12$\,rad/s; hold 6. \\
OpenLaptop & R & 600 & Open a laptop lid from $40$--$50^\circ$. & Lid reached $\geq 90^\circ$, then stays $\geq 85^\circ$; hold 4. \\
OpenPackage & B & 500 & Fold both flaps of a package open. & Both flaps rotated $> 90^\circ$ from closed. \\
\midrule
\multicolumn{5}{@{}l}{Grasping and placement} \\
\midrule
PickBottle & R & 900 & Take a bottle out of the cabinet and set it on a target. & Bottle out of the cabinet, upright within 0.15\,m of the target, at rest, released; hold 5. \\
PickCharger & R & 900 & Lift a charger out of the drawer and place it. & Charger out of the drawer, within 0.14\,m of the target, flat, at rest; hold 5. \\
PickBolt & L & 700 & Stand a bolt upright on its target. & Bolt head within 0.10\,m of the target, tilt $\leq 5^\circ$, at rest, released; hold 3. \\
PlaceBread & R & 400 & Move the bread onto the plate. & Bread within 0.13\,m of the plate centre and 0--7\,cm above it; hold 20. \\
ExtractApple & R & 700 & Take the apple out of the package and place it on the plate. & Apple lifted clear of the package, then on the plate within 5\,cm, at rest, hand $> 0.10$\,m away; hold 40. \\
AppleToBlender & R & 700 & Take the apple from the fridge into the blender pitcher. & Apple inside the pitcher and at rest, after leaving the fridge and being lifted $\geq 8.5$\,cm; hold 12. \\
LayFlatPanel & L & 600 & Take a panel from its rack and lay it flat in the placement zone. & Panel flat (normal $z \geq 0.95$) inside the zone, yaw $\leq 20^\circ$, at rest, released; hold 5. \\
PickDebris & R & 900 & Put three debris pieces into the dustpan. & All three pieces at rest inside the dustpan and none held; hold 8. \\
\midrule
\multicolumn{5}{@{}l}{Insertion and assembly} \\
\midrule
InsertPlug & R & 900 & Insert the charger plug into the socket. & Both pins 18--31\,mm deep, lateral offset $< 9$\,mm, tilt $\leq 18^\circ$, at rest; hold 5. \\
InsertRail & R & 700 & Slide a rail into the groove of the panel. & Rail seated: depth error $\leq 2$\,mm, lateral $\leq 4$\,mm, aligned, released, at rest; hold 5. \\
SeatNut & R & 800 & Place the nut onto the bolt. & Nut on the bolt axis (xy $< 5$\,mm), at seat height within 3\,mm, tilt $\leq 12^\circ$; hold 3. \\
PlaceLid & R & 400 & Put the lid on the blender pitcher. & Lid lifted $\geq 3$\,cm once, then seated: xy $\leq 5$\,mm, z $\leq 8$\,mm, tilt $\leq 15^\circ$; hold 15. \\
\midrule
\multicolumn{5}{@{}l}{Twisting and screwing} \\
\midrule
UnscrewCap & R & 1400 & Unscrew the bottle cap and set it down. & Cap rotated $\geq 180^\circ$ and detached, then flat within 0.10\,m of the target, at rest, hand $\geq 65$\,mm away; hold 5. \\
TurnNut & R & 900 & Tighten the seated nut on the bolt. & Net nut rotation $\geq 180^\circ$ while seated; hold 5. \\
FastenScrew & R & 600 & Turn a wing screw on the rail to its stop. & Screw angle $\geq 87^\circ$. \\
\midrule
\multicolumn{5}{@{}l}{Pressing} \\
\midrule
PressButton & R & 400 & Press the blender's power button. & Button pressed $\geq 80\%$ of its 2\,mm travel; hold 4. \\
ClickMouse & R & 500 & Click the left mouse button. & Left button $\geq 60\%$ travel with a finger pad over it, right button $< 35\%$; hold 3. \\
TypeKeys & L & 700 & Type a fixed five-key sequence ending with Enter. & Each key in turn pressed $\geq 60\%$ with a pad on the keycap, held 4 steps, then released. \\
AspirateLiquid & R & 800 & Draw liquid from the sample cup with the pipette. & Plunger pressed $\geq 7.5$\,mm with the tip in the cup, then released; pipette back in its stand; hand away. \\
DispenseLiquid & L & 1100 & Dispense from the pipette into the test tube. & Plunger pressed $\geq 7.5$\,mm with the tip in the tube, then released; pipette back in its stand; hand away. \\
\midrule
\multicolumn{5}{@{}l}{Scooping and pouring} \\
\midrule
ScoopTomato & R & 1200 & Scoop the tomato with the spoon onto the plate. & Tomato carried on the spoon, then on the plate within 9.4\,cm and 0--6\,cm above it; hold 20. \\
PourBeads & R & 700 & Pour the beads from the bottle into the bowl. & $\geq 12$ of 16 beads in the bowl, $\leq 2$ spilled, none airborne; hold 10. \\
DumpDustpan & R & 600 & Empty the dustpan into the bin. & All three debris pieces inside the bin; hold 10. \\
\end{xltabular}
\normalsize

\subsection{Parameters of the Environment}
\label{subsec:parameters_environment}
Environment parameters are organized into four explicit layers, as shown in Table~\ref{tab:layers}.
\begin{table}[t]
\centering\small
\caption{The four parameter layers of a DexArena environment. The first three are shared by every task in the benchmark. The fourth belongs to the task
designer.}
\label{tab:layers}
\setlength{\tabcolsep}{5pt}
\begin{tabularx}{\linewidth}{@{}ll>{\raggedright\arraybackslash}X@{}}
\toprule
Layer & Scope & Contents \\
\midrule
Environment & benchmark-wide &
Physics rate 400\,Hz and control rate 20\,Hz. Solver iterations 24/8, contact offset 1\,mm, deterministic solver. Work table geometry and friction. \\
\addlinespace
Hand & benchmark-wide &
Joint stiffness, armature and joint friction from the vendor model, unscaled; damping from the model's damping ratio; torque limits from the URDF. Finger-pad contact model ($\mu = 1.2$, 5\,mm torsional patch radius). \\
\addlinespace
Arm & benchmark-wide &
Absolute position of the platform and of the arm roots relative to the work table. Arm start configuration. \\
\addlinespace
Task & per task &
Objects, their contact materials, masses and inertias. Layout and its randomization ranges. Success conditions and hold durations. Per-episode
initialization. \\
\bottomrule
\end{tabularx}
\end{table}

\Cref{tab:layer_values} gives the values the first three layers are fixed to. The hand layer is shown for the Sharpa Hand, the hand used for all released data; the other supported hands follow the same layer with their own vendor values.

\begin{table}[t]
\centering\small
\caption{Values of the three fixed layers. Hand values are for the Sharpa Hand.}
\label{tab:layer_values}
\setlength{\tabcolsep}{4pt}
\begin{tabularx}{\linewidth}{@{}l>{\raggedright\arraybackslash}X@{}}
\toprule
\multicolumn{2}{@{}l}{Environment layer} \\
\midrule
Rates & Physics 400\,Hz, control 20\,Hz (20 physics sub-steps per control step) \\
Solver & TGS, 24 position and 8 velocity iterations, PCM, friction every iteration \\
Contact & Contact offset 1\,mm, rest offset 0, bounce threshold 2\,m/s \\
Integration & Enhanced determinism on, sleeping disabled, CCD off, gravity 9.81\,m/s$^2$ \\
Materials & Fallback $\mu=0.30$; table and ground $\mu=0.30$; restitution 0 throughout \\
\midrule
\multicolumn{2}{@{}l}{Hand layer (Sharpa Hand, 22 joints)} \\
\midrule
Joint drives & Stiffness 0.9--13.2\,N\,m/rad, armature and Coulomb joint friction as in the vendor model, unscaled; damping 0.02--0.37\,N\,m\,s/rad from the model's damping ratio of 0.9 \\
Torque limits & 0.19--3.3\,N\,m per joint, from the URDF effort limits \\
Finger pads & Pads and distal phalanges: $\mu=1.2$, restitution 0, torsional patch radius 5\,mm (2\,mm minimum) \\
Other hand links & $\mu=0.35$; self-collision enabled \\
Mount & 90$^\circ$ adapter between the arm flange and the hand \\
\midrule
\multicolumn{2}{@{}l}{Arm layer (Franka Panda)} \\
\midrule
Placement & Platform root at $x=-0.870$\,m, work-table centre at $x=+0.180$\,m, arm roots at $y=\pm0.220$\,m (bimanual), all in world coordinates \\
Start configuration & $(-0.50, -0.20, 0.00, -2.60, 0.00, 2.40, 0.79)$\,rad, plus per-episode noise from the task layer \\
Control & Joint impedance: stiffness 1000\,N\,m/rad, damping 100\,N\,m\,s/rad, torque limit 100\,N\,m, gravity compensation on; actions are absolute joint targets \\
Materials & Arm links $\mu=0.30$, restitution 0 \\
\bottomrule
\end{tabularx}
\end{table}

The task layer is the only layer a new task writes. It has to specify:
\begin{itemize}\setlength{\itemsep}{1pt}
\item the objects, with explicit mass and inertia consistent with the collision geometry; manipulated objects carry a contact material chosen by material class from a fixed table, fixtures use the default material; free bodies receive a velocity ceiling of 3\,m/s and a depenetration ceiling of 0.5\,m/s;
\item the layout: one set of nominal poses shared by the long-horizon task and its stages, within 0.54--0.90\,m of the arm base, and the per-episode randomization ranges, drawn from the episode's random stream together with the arm-start noise ($\sigma=0.02$\,rad per joint);
\item the success conditions of each stage with their hold durations, the episode horizon, and the stage flags kept in the scene state;
\item optional on-table guidance cues for the operator, which are visual only.
\end{itemize}

\subsection{Teleoperation System}
\label{subsec:teleop_system}

\textbf{Hardware.} The operator wears a Quest~3 headset and a pair of MANUS Meta Gloves Pro. A Quest~3 controller is mounted on the back of each hand on a custom bracket and supplies the wrist pose. The whole system, simulation included, runs on one workstation with a single RTX~5090.

\textbf{Retargeting.} The glove reports a 25-node hand skeleton. For the Sharpa Hand it is passed to the hand vendor's retargeting solver \citep{sharpa2026manussdk}, which returns a target angle for each of the 22 finger joints; the simulated hand is the kinematic model the solver was built for, so the angles are applied by joint name without rescaling. The other supported hands are retargeted with the open-source \texttt{dex-retargeting} library \citep{qin2023anyteleop}, with one configuration per hand. Both paths match finger kinematics only. 
% Grip force is whatever the joint servos produce at the resulting penetration, and the operator cannot command it directly.

\textbf{Wrist and arm.} The controller pose is passed through a critically damped second-order reference shaper and a position servo. The servo detects tracking dropouts, which the headset reports as a frozen position with the tracking flag still set, and holds the last valid target instead of following the frozen one. The wrist target is solved into seven absolute arm joint targets by differential inverse kinematics with a null-space term that pulls the redundant degree of freedom towards a comfort configuration; the usual end-effector delta interface carries no such term, and over a long session the redundancy winds up against the joint limits. Per-step joint increments are clamped to the real Franka velocity limits, so every recorded action is kinematically executable on the real arm. Bimanual tasks run the same chain
for each arm.

\textbf{Rendering for the operator.} The headset view is produced by a separate process that holds a render-only replica of the environment and is driven by the poses and joint positions published by the control process. It renders at 72\,fps, the headset's frame rate, independently of the 20\,Hz control loop.

\textbf{Operator aids.} The operator can freeze the fingers at their current targets while repositioning the wrist. On-table markers show where an object should be placed; they carry no collision geometry and are hidden in all released camera views.

\textbf{Segmentation.} An episode is armed by the operator and cut automatically 20 control steps (1\,s) after the success condition first holds, so every demonstration ends in a settled state. Four operators collected the data over about two weeks.

\subsection{Data Post-processing, Validation and Release}
\label{subsec:data_post-processing_validation_release}

\textbf{Review and export.} Recorded episodes are staged per session and reviewed afterwards; only episodes that end in success are exported. The first three frames of each episode are dropped at export, as they precede the operator's first command. Each exported episode is a LeRobot episode with the per-frame columns of \Cref{tab:columns}; the scene state of every frame is kept next to the dataset in the form the simulator restores from.

\begin{table}[h]
\centering\small
\caption{Per-frame columns of a released episode.}
\label{tab:columns}
\begin{tabularx}{\linewidth}{@{}l>{\raggedright\arraybackslash}X@{}}
\toprule
Column & Content \\
\midrule
\texttt{observation.state} & Robot joint positions and velocities followed by task-specific quantities (object poses, stage flags); dimension per stage in \Cref{tab:dims} \\
\texttt{action} & Absolute joint targets, 7 arm + 22 hand per arm (29 / 58) \\
\texttt{next.reward} & 1 while the success condition holds, else 0 \\
\texttt{next.success}, \texttt{next.done} & Success and episode-end flags \\
\texttt{observation.environment\_state} & Flattened poses and joint states of all scene bodies \\
\bottomrule
\end{tabularx}
\end{table}

\begin{table}[h]
\centering\footnotesize
\caption{Observation dimension of each stage. The action dimension is 29 for single-arm and 58 for bimanual stage-level tasks.}
\label{tab:dims}
\setlength{\tabcolsep}{3.5pt}
\begin{tabular}{@{}lr lr lr@{}}
\toprule
Stages & Obs & Stages & Obs & Stages & Obs \\
\midrule
OpenLunchBox & 85 & PlaceBread & 116 & UnscrewCap & 96 \\
OpenCabinet & 79 & ExtractApple & 85 & TurnNut & 97 \\
OpenDrawer & 79 & AppleToBlender & 88 & FastenScrew & 93 \\
OpenFridge & 74 & LayFlatPanel & 134 & PressButton & 111 \\
CloseFridge & 74 & PickDebris & 88 & ClickMouse & 149 \\
OpenLaptop & 123 & InsertPlug & 96 & TypeKeys & 390 \\
OpenPackage & 128 & InsertRail & 153 & AspirateLiquid & 83 \\
PickBottle & 97 & SeatNut & 95 & DispenseLiquid & 90 \\
PickCharger & 92 & PlaceLid & 121 & ScoopTomato & 118 \\
PickBolt & 82 & & & PourBeads & 92 \\
 & & & & DumpDustpan & 90 \\
\bottomrule
\end{tabular}
\end{table}

\textbf{Restoration check.} For sampled episodes of every dataset, recorded frames are restored in a fresh simulator instance and compared with the recording: the scene state, the observation, the task's own quantities and the rendered images must match. Quantities that depend on the solver's contact state cannot be reproduced from a restored frame and are excluded. A dataset is released only after this check passes.

\textbf{Open-loop replay.} The recorded actions of every accepted episode are replayed from the recorded initial state in a fresh simulator instance, and the episode counts as replayable if the success condition is reached.
The solver's contact state cannot be restored, so a replay starts from a slightly different contact state even when the recorded scene state is restored.
During contact-rich motion, such differences can grow, whereas a human operator would correct them in the loop.

\textbf{Release.} For each stage the released 100 demonstrations are the replayable episodes.

\textbf{Camera views.} Camera images are rendered offline by restoring each frame and rendering a fixed scene view and one wrist view per arm in use.
The scene camera is placed 1.08\,m above the table behind the workspace with its image plane tilted 20$^\circ$ from the tabletop, aimed at the centre of the task's scene content.
Each wrist camera sits on the wrist link with calibrated extrinsics.
All views are rendered at 256$\times$256 with a 90$^\circ$ vertical field of view.
Guidance markers are hidden.
Videos are encoded with AV1, and each released video is decoded and compared with the rendered frames (PSNR $\geq 30$\,dB) before release,  so that the released video is a faithful copy of the rendering, with no dropped or misaligned frames.

\subsection{Policy Implementations and Training Configurations}
\label{subsec:policy_implementations_training_configurations}
\paragraph{Proximal Policy Optimization (PPO).}
At step $t$ the agent receives an observation $o_t$.
It consists of two RGB images from fixed cameras and a proprioceptive vector $p_t$.
$p_t$ holds only quantities that are also available on the real robot: joint positions and velocities and the palm pose (TCP).
Privileged simulator states are explicitly excluded.
$n(\cdot)$ denotes standardization of $p_t$ with running estimates of its mean and variance.
Each image encoder uses one convolutional trunk per camera, followed by a linear projection, LayerNorm and $\tanh$; the per-camera outputs are concatenated into a latent vector.
The actor and the critic own separate encoders $f_a$ and $f_v$, each trained only by its own loss.
The action $a_t\in\mathbb R^{|A|}$ is the vector of absolute joint targets of the arm--hand system ($|A|=29$).
The policy is a Gaussian with a bounded mean and a state-independent scale:
\begin{equation}
  \pi_\psi(a_t\mid o_t)=\mathcal N\bigl(a_t;\;\mu_\psi(o_t),\,\operatorname{diag}\sigma^2\bigr),\qquad
  \mu_\psi(o_t)=b+c\odot\tanh\bigl(g_\psi([f_a(o_t),\,n(p_t)])\bigr).
\end{equation}
 $g_\psi$ represents an MLP with parameters $\psi$. $\sigma\in\mathbb R^{|A|}$ represents a learned standard-deviation vector that does not depend on $o_t$. $b$ and $c$ are the per-joint center and half-width of the interval that bounds the mean. This interval is the demonstrations' action range, padded by a small margin and intersected with the joint limits.

The output bias of $g_\psi$ is initialized so that the untrained policy outputs the mean demonstration action, and $\sigma$ is initialized to a small value.
The critic $V_\omega(o_t)$ is an MLP applied to $[f_v(o_t),\,n(p_t)]$.
Let $\mathrm{succ}_t\in\{0,1\}$ be the simulator's success flag.
The reward is this flag itself, $r_t=\mathrm{succ}_t$.
The episode terminates on success; reaching the time limit is a truncation, not a termination, so $V_\omega\in[0,1]$.
Advantages $\hat A_t$ are computed with generalized advantage estimation (discount $\gamma$, trace parameter $\lambda$), and the value targets are $\hat R_t=\hat A_t+V_\omega(o_t)$.
Bootstrapping happens only at truncations, using $V_\omega(o^{\mathrm{fin}}_t)$, where $o^{\mathrm{fin}}_t$ is the true final observation before the automatic reset.
A successful step contributes exactly $r_t$.

Demonstrations $\mathcal D_{\mathrm{demo}}=\{(o^d,a^d)\}$ carry no behavior log-probabilities, so they cannot enter the on-policy objective directly.
They are used in three other ways.
First, before any rollout the actor is warmed-up by minimizing the BC loss
$\mathcal L_{\mathrm{BC}}=\mathbb E_{\mathcal D_{\mathrm{demo}}}\bigl\|\mu_\psi(\operatorname{aug}(o^d))-a^d\bigr\|_2^2$,
where $\operatorname{aug}(\cdot)$ is a random image shift.
During this warm-up, the proprioceptive statistics are initialized from the demonstrations.
Second, $\mathcal L_{\mathrm{BC}}$ remains as an anchor during training.
Third, environments are reset to demonstration states under a reverse curriculum.
Let $q\in[0,1]$ be the normalized progress of a frame within its demonstration.
At each reset, with probability $p(\tau_c)$ the simulator is restored to the physical state of a demonstration frame with $q\ge 1-f(\tau_c)$.
Here $\tau_c$ is a curriculum clock that advances only while the recent training success rate exceeds a threshold.
As $\tau_c$ advances, $p$ decays and $f$ grows linearly.
Early episodes therefore start close to success, and later episodes increasingly start from the task's own initial-state distribution.
Each update runs over mini-batches of the latest rollout and minimizes
\begin{equation}
  \mathcal L=c_v\,\mathcal L_V+\kappa\bigl[\mathcal L_{\mathrm{clip}}+\beta(t)\,\mathcal L_{\mathrm{BC}}\bigr],\qquad
  \mathcal L_{\mathrm{clip}}=\mathbb E\Bigl[\max\bigl(-\bar A\,\rho,\;-\bar A\,\operatorname{clip}(\rho,1-\epsilon,1+\epsilon)\bigr)\Bigr].
\end{equation}
$\rho=\pi_\psi(a\mid o_t)/\pi_{\mathrm{old}}(a\mid o_t)$ is the probability ratio between the current policy and the policy $\pi_{\mathrm{old}}$ that collected the rollout. $\bar A$ is $\hat A$ standardized within the mini-batch, and $\epsilon$ is the clipping range. $\mathcal L_V=\tfrac12\mathbb E\bigl[(V_\omega(\operatorname{aug}(o_t))-\hat R)^2\bigr]$ is the value loss, weighted by $c_v$. $\beta(t)$ is the BC weight. It decays linearly with the environment step $t$ to a small floor. $\kappa\in\{0,1\}$ is a KL gate. It equals 1 until the approximate KL divergence between $\pi_\psi$ and $\pi_{\mathrm{old}}$ on some mini-batch exceeds a threshold, and 0 for the rest of that update. After that point the actor and the BC anchor are frozen, while the value function keeps training.
No entropy bonus is used.
$\rho$ is evaluated on unaugmented frames, and $\log\pi_{\mathrm{old}}$ is recomputed at the start of each update, so that both policies are compared on the same input.
Augmentation is applied only to the value and BC branches.
The proprioceptive statistics are held fixed during each rollout and update and refreshed afterwards.
The actor and critic parameters use separate learning rates.

\paragraph{Reinforcement Learning with Prior Data (RLPD).}
RLPD is built on soft actor-critic (SAC), an off-policy method that trains a stochastic policy
to maximize the return plus an entropy bonus.
Relative to SAC, RLPD makes three changes.
First, it uses symmetric sampling: every mini-batch is drawn half from the demonstration dataset
$\mathcal D_{\mathrm{demo}}$ and half from the online replay buffer $\mathcal D_{\mathrm{replay}}$, and it applies LayerNorm to efficiently prevent overestimation.
Second, it replaces SAC's two critics with an ensemble of LayerNorm critics, and its target takes
the minimum over a random subset of heads.
Third, it allows several gradient updates per environment step.
We use RLPD without algorithmic changes and adapt only the input and action pipeline.

Observations and image encoders are the same as PPO.
However, a single encoder $f_\phi$ with parameters $\phi$ is used, giving the state feature
$s_t=[f_\phi(o_t),\,n(p_t)]$.
The policy does not output absolute joint targets directly.
Instead it outputs a bounded increment $u_t\in[-1,1]^{|A|}$, and the command sent to the robot is
\begin{equation}
  a_t=\operatorname{clip}\bigl(a_{t-1}+\delta\,u_t,\;a_{\mathrm{low}},\,a_{\mathrm{high}}\bigr).
\end{equation}
Here $\delta$ is the maximum per-step change in radians, and $[a_{\mathrm{low}},a_{\mathrm{high}}]$
are the joint limits.
The previous command $a_{t-1}$ is appended to $p_t$ so that the state remains Markovian.
At $t=0$, $a_{-1}$ is set to the mean first-frame demonstration action.
Both the actor and the critic operate on the increment $u_t$.
Demonstration actions are converted to the same increments before training.
The actor $\pi_\psi(u\mid s)$ is a tanh-squashed Gaussian whose mean and standard deviation both
depend on $s$.
The critic is an ensemble of $N$ Q-functions $Q_{\theta_k}(s,u)$, $k=1,\dots,N$, each an MLP with
LayerNorm.
Each $Q_{\theta_k}$ has a target copy $Q_{\bar\theta_k}$.
The reward is the success flag shifted by $-1$, $\tilde r_t=\mathrm{succ}_t-1\in\{-1,0\}$, a per-step
cost that stops on success.
The termination flag $d_t$ equals $\mathrm{succ}_t$.
Time-limit truncations are therefore not terminal and are bootstrapped.
Under this reward, $Q$ equals the negative discounted number of steps to success.

Each mini-batch $\mathcal B$ contains $B$ transitions $(o_t,u_t,\tilde r_t,d_t,o_{t+1})$,
half from each source, and both halves are treated identically.
The critic is trained as in SAC:
\begin{equation}
  \mathcal L_Q=\frac{1}{NB}\sum_{k=1}^{N}\sum_{t\in\mathcal B}\bigl(Q_{\theta_k}(s_t,u_t)-y_t\bigr)^2,\qquad
  y_t=\tilde r_t+\gamma\,(1-d_t)\min_{k\in\mathcal K}Q_{\bar\theta_k}(s_{t+1},u'),
\end{equation}
where $u'\sim\pi_\psi(\cdot\mid s_{t+1})$ and $\mathcal K$ is a random subset of $M<N$ heads.
Unlike SAC, the target contains no entropy term.
The next-state feature $s_{t+1}$ is computed by the current encoder under stop-gradient
$\operatorname{sg}(\cdot)$, which blocks gradients from flowing back through it.
Hence $y_t$ serves as a fixed regression target, and no separate target encoder is kept.
$\mathcal L_Q$ is the only loss that trains the encoder.
The actor receives the detached feature $\bar s_t=[\operatorname{sg}(f_\phi(o_t)),\,n(p_t)]$ and minimizes
\begin{equation}
  \mathcal L_\pi=\frac1B\sum_{t\in\mathcal B}\Bigl[\alpha\log\pi_\psi(\tilde u_t\mid\bar s_t)
  -\frac1N\sum_{k=1}^{N}Q_{\theta_k}(\bar s_t,\tilde u_t)\Bigr],
\end{equation}
where $\tilde u_t$ is a reparameterized sample from $\pi_\psi(\cdot\mid\bar s_t)$.
The temperature $\alpha$ is tuned automatically.
The target critics track the online ones by Polyak averaging.

\begin{figure}
    \centering
    \includegraphics[width=\textwidth]{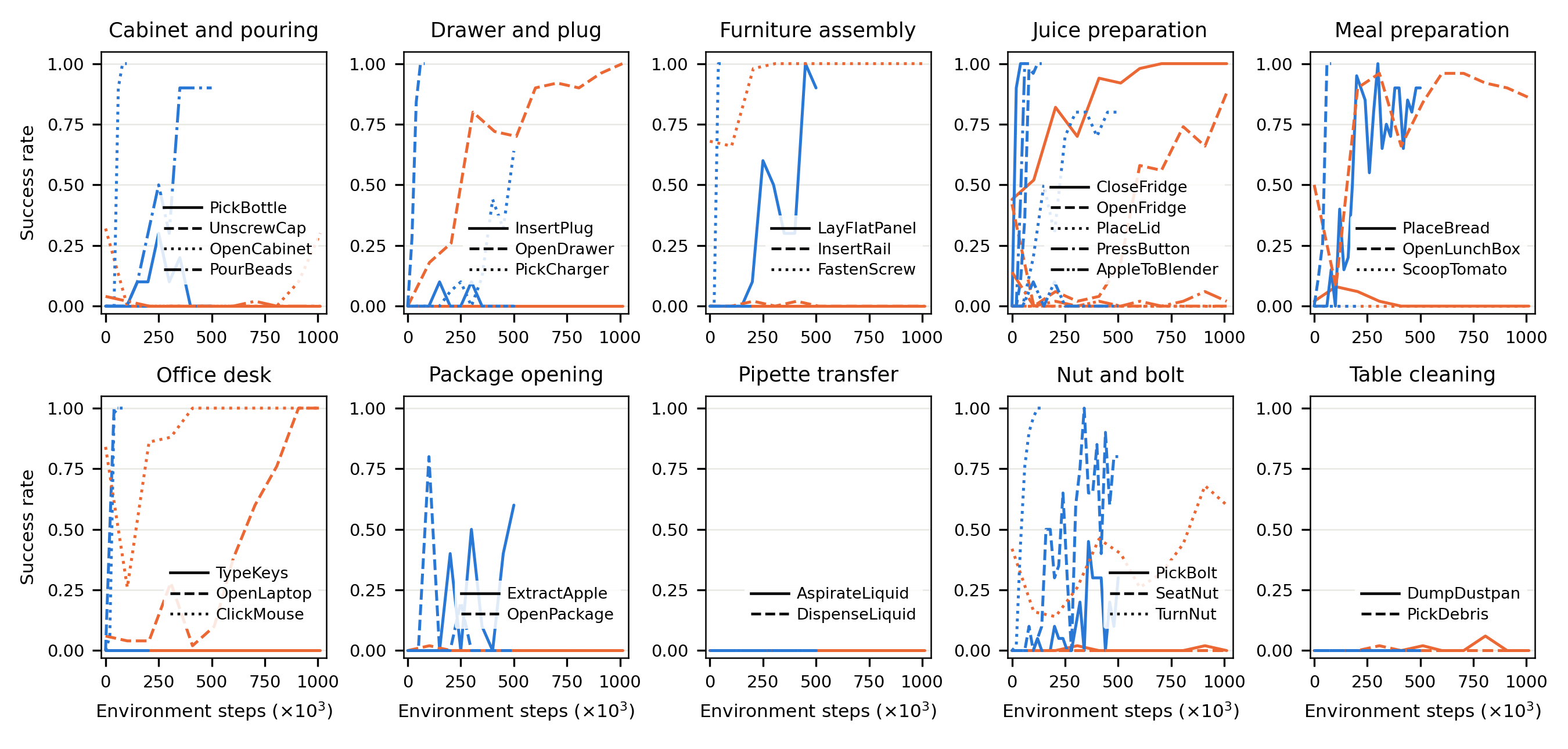}
    \caption{Training curves of the RL baselines on the 30 stage-level tasks, grouped by long-horizon task.
    Orange curves are PPO and blue curves are RLPD; line styles distinguish the stage-level tasks within each long-horizon task.
    The point at 0 steps for PPO is the policy after BC warm-up.
    PPO is trained for a fixed budget of $10^{6}$ environment steps on every stage-level task.
    RLPD is trained for up to $5\times10^{5}$ environment steps and stopped early once the evaluation success rate reaches 1.0 in two consecutive evaluations or stays at 0 for ten consecutive evaluations, so each RLPD curve ends where its run was stopped.}
    \label{fig:rl_training_curves}
\end{figure}

\paragraph{Action Chunking with Transformers (ACT).}
The visual observation of ACT consists of $256\times256$ RGB images from a fixed front camera and a task-relevant wrist camera. For each camera, we use the frames at temporal offsets $\{0,2,4\}$, resulting in six visual inputs per policy query. The proprioceptive input contains only the current robot joint positions; no privileged object state is provided. Both methods output absolute target joint positions under joint-position control. The state and action dimensions are 29 for single-arm tasks, comprising 7 arm joints and 22 hand joints, and 58 for bimanual tasks.
A shared ImageNet-pretrained ResNet-18 converts each camera image into spatial feature tokens. The image tokens and an encoded proprioceptive-state token are fused by a Transformer to obtain the observation context. A conditional-VAE posterior encoder takes the context and demonstrated action sequence as input, while the decoder predicts a 25-step action chunk.

Let $\hat{\mathbf{a}}^{\mathrm{post}}$ and
$\hat{\mathbf{a}}^{0}$ denote the action chunks decoded from the sampled
posterior latent and the zero latent, respectively. The training objective is
\begin{equation}
    \mathcal{L}_{\mathrm{ACT}}
    =
    \mathcal{L}_{\mathrm{rec}}
    \left(\hat{\mathbf{a}}^{\mathrm{post}},\mathbf{a}\right)
    +
    \lambda_{0}
    \mathcal{L}_{\mathrm{rec}}
    \left(\hat{\mathbf{a}}^{0},\mathbf{a}\right)
    +
    \lambda_{\mathrm{KL}}
    D_{\mathrm{KL}}
    \left(q(\mathbf{z}\mid\mathbf{o},\mathbf{a})
    \,\|\,\mathcal{N}(\mathbf{0},\mathbf{I})\right),
\end{equation}
where $\mathcal{L}_{\mathrm{rec}}$ is a masked $\ell_1$ loss that excludes
padded actions. We compute the reconstruction loss separately for the arm and
hand action groups, using weights of 2 and 1, respectively. For bimanual tasks,
the same weighting is applied independently to the two arms and hands. We use
$\lambda_{0}=1$ and $\lambda_{\mathrm{KL}}=10^{-3}$. The zero-latent term
directly trains the inference path instead of relying only on the sampled
posterior latent.

At inference time, the latent variable is fixed to zero and the policy predicts
a new action chunk at every control step. All predictions corresponding to the
current step are combined using temporal ensembling. A prediction generated
$\Delta t$ steps earlier is assigned weight
\begin{equation}
    w(\Delta t) \propto \exp(-\alpha \Delta t),
\end{equation}
where $\alpha=0.05$. This allows the controller to combine the temporal
consistency of previous predictions with the responsiveness of newly generated
commands.

\paragraph{Diffusion Policy (DP).}
Our DP implementation uses the same visual history, proprioceptive state, and
absolute joint-position action representation as ACT. A shared
ImageNet-pretrained ResNet-18 with Group Normalization encodes each image into a
128-dimensional feature. The features from all camera and temporal inputs are
concatenated with the current joint positions and used as the global condition
of a temporal U-Net. The model generates a 16-step action sequence.

During training, the actions are normalized to $[-1,1]$ using the per-dimension
minimum and maximum values computed from the training split. Given a clean
action sequence $\mathbf{a}^{0}$, a diffusion timestep $k$, and Gaussian noise
$\boldsymbol{\epsilon}$, the noisy action sequence is
\begin{equation}
    \mathbf{a}^{k}
    =
    \sqrt{\bar{\alpha}_{k}}\,\mathbf{a}^{0}
    +
    \sqrt{1-\bar{\alpha}_{k}}\,\boldsymbol{\epsilon}.
\end{equation}
The temporal U-Net is trained to predict the added noise:
\begin{equation}
    \mathcal{L}_{\mathrm{DP}}
    =
    \mathbb{E}_{k,\boldsymbol{\epsilon}}
    \left[
        \left\|
        \boldsymbol{\epsilon}
        -
        \boldsymbol{\epsilon}_{\theta}
        \left(\mathbf{a}^{k},k,\mathbf{o}\right)
        \right\|_{2}^{2}
    \right].
\end{equation}
Losses on padded action steps are masked out. We maintain an exponential moving
average of both the visual encoder and the temporal U-Net and use the averaged
model for evaluation.

At inference time, a 16-step action sequence is initialized from Gaussian noise
and generated using 100 reverse DDPM steps with a cosine noise schedule. The
visual observations are encoded once for each generated chunk. The controller
executes the first four actions and then generates a new sequence conditioned on
the latest observations. Therefore, DP operates as a receding-horizon controller
with an execution horizon of four steps.

\begin{table}[t]
    \centering
    \caption{Training configurations of PPO and RLPD. Both methods use the
    same observation interface (two $128\times128$ RGB cameras plus
    proprioception) and the same demonstrations (100 VR teleoperation episodes
    per stage-level task). ``--'' means not applicable.}
    \label{tab:rl_config}
    \small
    \setlength{\tabcolsep}{5pt}
    \begin{tabularx}{\textwidth}{lXX}
        \toprule
        \textbf{Configuration} & \textbf{PPO} & \textbf{RLPD} \\
        \midrule
        \multicolumn{3}{l}{\textit{Observations and actions}} \\
        Image resolution
        & $128\times128$, base and wrist camera & $128\times128$, base and wrist camera \\
        Proprioception
        & $q$, $\dot q$, EE pose & $q$, $\dot q$, EE pose, $a_{t-1}$ \\
        Action
        & Absolute joint targets, 29-D & Joint-target increments, 29-D \\
        Action bound
        & $\tanh$, demo range $+10\%$ & $0.2\,u_t$, $u_t\in[-1,1]$ \\
        Policy distribution
        & Gaussian, $\log\sigma_0=-3.5$ & $\tanh$-Gaussian, $\log\sigma\in[-5,2]$ \\
        \midrule
        \multicolumn{3}{l}{\textit{Networks}} \\
        Visual encoder
        & 4-layer CNN, from scratch (32-64-128-256) & 4-layer CNN, from scratch (32-64-128-256) \\
        Latent per camera
        & 128 & 64 \\
        Encoder gradient
        & Actor and critic & Critic only \\
        MLP hidden layers
        & $4\times512$ & $3\times256$ \\
        Critic ensemble
        & -- & 5 heads, min over 2 \\
        Image augmentation
        & Random shift, pad 4 & Random shift, pad 4 \\
        \midrule
        \multicolumn{3}{l}{\textit{Reward and objective}} \\
        Reward
        & $\mathbf{1}[\text{success}]$ & $\mathbf{1}[\text{success}]-1$ \\
        Discount $\gamma$
        & 0.99 & 0.999 \\
        GAE $\lambda$
        & 0.95 & -- \\
        Clip coefficient / target KL
        & 0.2 / 0.2 & -- \\
        Value / entropy coefficient
        & 0.5 / 0 & -- \\
        Entropy temperature
        & -- & Auto-tuned \\
        Target network $\tau$
        & -- & 0.005 \\
        \midrule
        \multicolumn{3}{l}{\textit{Optimization}} \\
        Optimizer
        & Adam & Adam \\
        Learning rate
        & $3\times10^{-5}$ actor, $3\times10^{-4}$ critic & $3\times10^{-4}$ \\
        Gradient clipping
        & 0.5 & 10 \\
        Parallel environments
        & 64 & 1 \\
        Batch size
        & $64\times200$, 32 minibatches & 256 (50\% demo) \\
        Epochs / UTD
        & 1 epoch & UTD 1 \\
        Replay buffer
        & -- & $2\times10^{5}$ \\
        Random warm-up
        & -- & 5{,}000 steps \\
        \midrule
        \multicolumn{3}{l}{\textit{Demonstration usage}} \\
        Demonstrations per stage-level task
        & 100 & 100 \\
        BC warm-up
        & 20{,}000 steps, batch 256 & -- \\
        BC anchor coefficient
        & $1.0\to0.05$ over $2.5\times10^{6}$ & -- \\
        Demo-state reset probability
        & $1.0\to0.1$ over $3.5\times10^{6}$ & -- \\
        Reverse-curriculum window
        & $20\%\to100\%$ over $2.5\times10^{6}$ & -- \\
        Curriculum pacing
        & $\times3$ / $\times1$ / $\times0.5$ at success $\ge0.8$ / $\ge0.5$ / $<0.5$ & -- \\
        Demo sampling
        & -- & 50\% of every batch \\
        \midrule
        \multicolumn{3}{l}{\textit{Budget and evaluation}} \\
        Environment steps
        & $1\times10^{6}$ & $5\times10^{5}$ \\
        Episode length
        & 500 & 500\\
        In-training evaluation
        & 50 episodes every $10^{5}$ steps & 20 episodes every 20--50k steps \\
        Reported evaluation
        & Best checkpoint, 50 episodes & Best checkpoint, 50 episodes \\
        Seed
        & 1 & 1 \\
        \bottomrule
    \end{tabularx}
\end{table}

\begin{table*}[t]
    \centering
    \caption{Implementation and training configurations of ACT and Diffusion
    Policy. Parameters shared by both methods are listed in the same row.}
    \label{tab:act_dp_config}
    \small
    \setlength{\tabcolsep}{5pt}
    \begin{tabularx}{\textwidth}{lXX}
        \toprule
        \textbf{Configuration} & \textbf{ACT} & \textbf{Diffusion Policy} \\
        \midrule
        Visual observations
        & Front and wrist RGB images at offsets $\{0,2,4\}$
        & Front and wrist RGB images at offsets $\{0,2,4\}$ \\

        Image resolution
        & $256\times256$
        & $256\times256$ \\

        Proprioceptive state
        & Current robot joint positions
        & Current robot joint positions \\

        Action representation
        & Absolute target joint positions
        & Absolute target joint positions \\

        State/action dimension
        & 29 (single-arm), 58 (bimanual)
        & 29 (single-arm), 58 (bimanual) \\

        Visual encoder
        & ImageNet-pretrained ResNet-18
        & ImageNet-pretrained ResNet-18 with GroupNorm \\

        Action normalization
        & Per-dimension mean and standard deviation
        & Per-dimension minimum and maximum \\

        Prediction horizon
        & 25 steps
        & 16 steps \\

        Hidden dimensions
        & Transformer hidden size 256; latent size 32
        & U-Net channels $(256,512,1024)$ \\

        Transformer configuration
        & 4 heads; 2 posterior, 2 fusion, and 4 decoder layers
        & -- \\

        Diffusion configuration
        & --
        & 100 DDPM steps; cosine noise schedule \\

        Reconstruction objective
        & Masked group-weighted $\ell_1$
        & Masked noise-prediction MSE \\

        Arm/hand loss weights
        & $2:1$ for each arm--hand pair
        & Uniform over action dimensions \\

        Zero-latent loss weight $\lambda_0$
        & 1.0
        & -- \\

        KL weight $\lambda_{\mathrm{KL}}$
        & $10^{-3}$
        & -- \\

        Batch size
        & 16
        & 16 \\

        Training updates
        & 20,000
        & 20,000 \\

        Optimizer
        & AdamW
        & AdamW \\

        Training random seed
        & 1
        & 1 \\

        Learning rate
        & $10^{-4}$
        & $10^{-4}$ \\

        Weight decay
        & $10^{-4}$
        & $10^{-4}$ \\

        Learning-rate schedule
        & Constant
        & 500-step warm-up followed by cosine decay \\

        Gradient clipping
        & Maximum norm 10
        & Maximum norm 10 \\

        Train/validation split
        & 90\%/10\% by episode
        & 90\%/10\% by episode \\

        Inference model
        & Zero-latent prediction
        & Exponential-moving-average model \\

        Inference strategy
        & Temporal ensemble at every control step
        & Execute 4 actions and replan \\

        Temporal-ensemble decay
        & 0.05
        & -- \\

        Checkpoint interval
        & 1,000 updates
        & 1,000 updates \\

        Stage-level evaluation seeds
        & $50000$--$50049$
        & $50000$--$50049$ \\
        \bottomrule
    \end{tabularx}
\end{table*}

% BEGIN PI05 POOLED LONG-HORIZON PROTOCOL 2026-10-16
\FloatBarrier
\paragraph{$\pi_{0.5}$ (LoRA).}
\label{app:pi05_pooled_protocol}
\textbf{Observation and action adaptation.}
Both standalone stage-level and long-horizon policies use LoRA. Their shared interface aligns external/wrist RGB with explicitly extracted robot joints. One wrist view is used per active arm; images are resized to $224\times224$ and missing views are masked. Proprioceptive features and controller actions are mapped separately by joint name. Physical state/action widths are 29 (seven arm and 22 hand joints) and 58 for single-arm and bimanual control, respectively; model capacities are 32, including padding, and 58. Bimanual action input/output projections expand from 32 to 58 channels, copying pretrained 32-channel slices while retaining initialized additional channels. Unlike GR00T's mixed arm/hand representation, all physical action channels predict absolute joint-position targets.

For pooled tasks containing single-arm stages, each 29-dimensional source is projected into the named left- or right-side slots of the complete 58-dimensional controller layout. Missing physical state channels use zero-valued \emph{normalized input placeholders}, not fabricated joint measurements; missing action channels and padded episode tails are excluded from supervision. Training-set 1st/99th percentiles are recomputed over the pooled observed channels and reused at inference. The canonical physical-width state is discretized into prompt tokens before model padding, so the three padded channels of a 29-joint single-arm task do not create extra state tokens. At deployment, only the stage's commanded channels are decoded; all other physical joints retain their previous actual absolute targets.

\textbf{Demonstrations and language conditioning.}
Each standalone stage-level model uses its task's 100 demonstrations, assigned evenly to the four official instruction phrasings (25 each) in a task-specific round-robin order. Each pooled long-horizon model uses 200--500 demonstrations across its stages and trains with each stage's official instruction index 0. Pooled sampling is frame-weighted, preserves episode boundaries and tail masks, and never concatenates stages. Both settings use reproducible official environment-ID/episode-seed instruction selection at evaluation. Data identities, normalization statistics and instruction versions are fixed within each run.

\textbf{Training.}
Each task-specific policy starts from the original OpenPI \texttt{pi05\_base} checkpoint and is trained with native JAX/Flax OpenPI building blocks at pinned source-fork revision \texttt{8d23b0f}. We use LoRA~\citep{hu2022lora} with masked supervised flow matching: VLM rank/alpha is $16/16$, and action-expert rank/alpha is $32/32$. The non-LoRA language/action-expert Transformer weights are frozen, while the LoRA matrices, visual encoder, action projections, and time-conditioning MLPs remain trainable.

All runs use global batch 32 and chunk length 30. The 29 single-arm standalone tasks use 30,000 optimizer updates each; bimanual OpenPackage uses 60,000. Pooled long-horizon budgets are 30,000 for TableClean and 60,000 for the other nine bimanual tasks. Drawer, Furniture and Office use microbatch $16\times2$ to retain batch 32. \Cref{tab:pi05_training_config} lists optimizer, precision, augmentation and learning-rate settings. Pooled segments restore parameters, optimizer, update count and RNG from the same run.

\textbf{Long-horizon models.}
Each long-horizon task uses one model trained on pooled stage demonstrations. A fixed final checkpoint controls the full native episode, with a single initial reset; stage transitions change only the instruction, not weights or simulator state. Success requires all stages to complete in order and native full-task success within the horizon, not an intersection of independently reset stage-level successes.

\textbf{Inference and evaluation.}
The policy predicts 30-step chunks with ten flow-sampling steps. The control loop runs at the benchmark's 20\,Hz rate and requests a new chunk asynchronously when fewer than 24 executable actions remain. Action timestamps refer to the observation's control step, not the response-arrival time; expired prefixes are discarded. In an overlap of $m$ executable predictions, the $k$th new prediction is blended with the queued target using weight $w_k=(k+1)/(m+1)$, for $k=0,\ldots,m-1$, in physical absolute-joint space. Targets are clipped to native joint limits. An empty queue holds the last actual target while control and termination checks continue. A stage transition clears the old action queue and invalidates stale responses, while preserving the global control step, held targets, single model, and live scene.

Formal evaluation uses the final target-budget checkpoint with 50 episodes (seeds 10001--10050); RNG keys fold in the episode seed and request generation/serial. Functional probes are separate; infrastructure errors are not policy failures. Acceptance requires 50 raw episodes, a final result and checkpoint/source provenance (Appendix~\ref{appendix:long_horizon_performance}).

\begin{table}[p]
\centering
\caption{\textbf{VLA training and evaluation configurations.} Settings for $\pi_{0.5}$-LoRA and GR00T N1.7.}
\label{tab:vla_training_config}
\begin{subtable}{\linewidth}
\centering
\caption{$\pi_{0.5}$ (LoRA): standalone stage-level and pooled long-horizon models.}
\label{tab:pi05_training_config}
\footnotesize
\setlength{\tabcolsep}{5pt}
\renewcommand{\arraystretch}{1.08}
\begin{tabular}{@{}p{0.35\linewidth}p{0.59\linewidth}@{}}
\toprule
\rowcolor{tablehead}
\textbf{Configuration} & \textbf{Setting} \\
\midrule
Pretrained checkpoint & OpenPI \texttt{pi05\_base}; original native parameters \\
Implementation & Native JAX/Flax OpenPI; source-fork revision \texttt{8d23b0f} \\
Training organization & Stage-level: one independent model per task; long-horizon: one pooled model per long task \\
Demonstrations per model & Stage-level: 100 from its own task; pooled: 200--500 across constituent stages \\
Training language & Stage-level: four official phrasings, 25 demonstrations each; pooled: official stage index 0 \\
Trainable modules & LoRA matrices, vision encoder, action projections and time-conditioning MLPs \\
Frozen modules & Non-LoRA language/action-expert Transformer parameters \\
VLM / action-expert LoRA & Rank/alpha $16/16$ and $32/32$, respectively \\
Physical / model dimensions & 29 / 32 (single arm); 58 / 58 (bimanual) \\
Action projection initialization & Copy pretrained 32-channel slices; initialize additional channels for 58-D \\
Visual inputs & External and active wrist RGB; unavailable views masked \\
Model image resolution & $224\times224$ per view \\
Training image augmentation & External view: $0.95$ crop and $\pm5^\circ$ rotation; all views: color jitter (0.3, 0.4, 0.5) \\
Evaluation augmentation & No random image augmentation \\
Pooled state/action normalization & Observed-channel 1st/99th percentiles \\
State encoding & Discrete physical-width state tokens before model padding \\
\midrule
Optimizer & AdamW; $\beta=(0.9,0.95)$; $\epsilon=10^{-8}$ \\
Global batch / accumulation & 32; microbatch $16\times2$ for Drawer, Furniture and Office \\
Stage-level target updates & 30,000 for each of 29 single-arm tasks; 60,000 for bimanual OpenPackage \\
Pooled target updates & 30,000 (TableClean); 60,000 (each of the other nine tasks) \\
Learning-rate schedule & 10,000-update warmup to $5\times10^{-5}$, then constant \\
Weight decay / clipping & $10^{-10}$ / global gradient norm 1.0 \\
Precision / EMA & BF16 model/frozen parameters; FP32 trainable leaves; EMA disabled \\
Training seed / token limit & 1000 / 250 tokens \\
Pooled segment continuation & At most 10,000 updates or 20 training hours; full optimizer/RNG restore \\
Hardware provenance & GPU model/count and memory recipe pinned in per-segment run receipts \\
\midrule
Evaluation language & Seeded official stage-set selection in both settings \\
Chunk / flow-sampling steps & 30 control steps / 10 \\
Control / asynchronous request & 20\,Hz; request when fewer than 24 executable actions remain \\
Overlap blending & Linear ramp $w_k=(k+1)/(m+1)$ over $m$ overlapping targets \\
Missing actions / empty queue & Hold previous actual absolute targets; native limits enforced \\
Long-horizon execution & One fixed checkpoint; no inter-stage scene reset or weight switching \\
Formal evaluation & 50 episodes per task; seeds 10001--10050; final target-budget checkpoint \\
\bottomrule
\end{tabular}
\end{subtable}
\end{table}
\FloatBarrier
% END PI05 POOLED LONG-HORIZON PROTOCOL 2026-10-16

\paragraph{GR00T N1.7.}
\textbf{Observation and action adaptation.}
We fine-tune through GR00T's own training entry point and evaluate the resulting
policy in process, inside the same closed-loop simulator used by every other
baseline. Single-arm tasks use one external and one wrist RGB view; the bimanual
task uses one external view and one wrist view per arm. Images are stored at
$256\times256$ and aligned in time with the corresponding robot states and action
targets. Images are re-rendered by restoring recorded simulator states; this path was checked against per-frame resets for pixel and state agreement.

We explicitly extract actuated robot joints from the raw observation and reorder articulation joints into controller order. Undriven joints raise an error rather than entering the state silently. Single-arm Sharpa control comprises
seven arm joints and 22 hand joints. Bimanual control comprises 58 targets ordered
by the controller's sub-controllers (left arm, right arm, left hand, right hand), not by concatenating two single-arm vectors. The action head
admits up to 132 dimensions, so the 29- and 58-dimensional layouts are used without
architectural widening; each task is registered as a new embodiment whose state and
action encoders are randomly initialized.

State and action vectors are declared as named joint groups rather than as one flat
vector, and each group carries its own action representation: arm groups are
predicted as offsets from the current configuration, hand groups as absolute
targets. Normalization statistics are computed from the training dataset and stored
in the dataset metadata alongside the group declaration. At inference, outputs are
inverse-transformed into absolute joint-position targets and clipped to the
controller's bounds.

\textbf{Demonstrations and language conditioning.}
Each stage-level task has 100 demonstrations, converted to the LeRobot v2.1 layout
with named joint groups in the metadata. Standalone stage-level models train and evaluate with one fixed stage instruction (e.g.\ ``open the cabinet door''). Code commits and instruction-table versions are fixed per experiment.

\textbf{Training.}
Each of the 30 stage-level tasks is fine-tuned independently from the pretrained
GR00T N1.7 checkpoint using its own demonstrations. We retain the flow-matching
objective and update all parameters, including the visual encoder and the language
backbone, without freezing any module and without LoRA. The published pooled-model processor configurations specify a $230\times230$ random training crop resized to $256\times256$, with color jitter and rotation disabled; evaluation uses the corresponding center crop. Each stage-level model uses 12,000 updates, global batch 32 (microbatch $2\times16$ accumulation), and seed 42. Segments resume optimizer state, schedule and step count. Checkpoints are saved every 1,000 updates; evaluation uses the final checkpoint, without evaluation-based selection. The configuration is summarized in
\Cref{tab:groot_training_config}.

\textbf{Long-horizon models.}
For the long-horizon evaluation we additionally fine-tune one policy per
long-horizon task on the pooled demonstrations of all its stage-level tasks (100
per stage, i.e.\ 200--500 episodes). Within each stage's pooled training data, four official instruction phrasings are assigned to 25 demonstrations each. Each pooled model uses 12,000 optimizer updates and a global batch size of 32, matching the update budget of one standalone stage-level model; equal update counts do not imply equal wall-clock compute. For \emph{Package opening}, whose two stages
use different embodiments (bimanual OpenPackage, right-arm ExtractApple), the pooled
dataset uses the 58-dimensional bimanual layout with the external and right-wrist
views; the single-arm episodes are padded with the bimanual robot's initial
left-arm configuration, and only the right-arm outputs are executed on that stage.
At evaluation the model runs one uninterrupted episode in the long-horizon
environment and the instruction is switched to the next stage as soon as the
environment reports the current stage complete.

\textbf{Inference and evaluation.}
The policy predicts a 40-step action chunk with four denoising steps and replans
from the latest observation after executing the first eight actions. Each
stage-level task is evaluated for 50 episodes using the final checkpoint under the
task's registered episode limit. Episode $i$ ($i=0,\ldots,49$) resets its environment with seed $10001+i$. These seeds are shared across stage-level tasks, generalization scenes and long tasks within their respective layouts; a long-task episode resets only once. The evaluation process seeds \texttt{random}, NumPy and PyTorch with 10001, also fixing the sampler's initial noise. The
instruction is the fixed training sentence for standalone stage-level models. For pooled long-horizon models, stage instructions are sampled reproducibly from the official sets under the registered seeded instruction-selection protocol. Episode seeds, episode lengths
and task outcomes are recorded. An episode counts as a success if the environment's
native success condition holds at any step.

\begin{table}[p]
\ContinuedFloat
\centering
\caption[]{\textbf{VLA training and evaluation configurations (continued).}}
\begin{subtable}{\linewidth}
\centering
\setcounter{subtable}{1}
\caption{GR00T N1.7: standalone stage-level and pooled long-horizon models.}
\label{tab:groot_training_config}
\footnotesize
\setlength{\tabcolsep}{5pt}
\renewcommand{\arraystretch}{1.08}
\begin{tabular}{@{}p{0.35\linewidth}p{0.59\linewidth}@{}}
\toprule
\rowcolor{tablehead}
\textbf{Configuration} & \textbf{Setting} \\
\midrule
Pretrained model & GR00T N1.7-3B \\
Training organization & 30 standalone stage policies and 10 pooled long-task policies \\
Demonstrations & 100 per stage-level task \\
Language instruction & Standalone: fixed stage sentence; pooled: four official phrases, 25 demos each per stage, with seeded official-set selection at evaluation \\
Fine-tuning & All parameters \\
Training objective & Flow matching \\
Vision encoder & Trainable \\
LoRA & Disabled \\
Embodiment & New embodiment; state and action encoders randomly initialized \\
Single-arm visual inputs & External RGB and wrist RGB \\
Bimanual visual inputs & External RGB and left/right wrist RGB \\
Stored image resolution & $256 \times 256$ per view \\
Proprioception & Joint positions in controller order \\
Physical state/action dimensions & 29 (single-arm); 58 (bimanual) \\
Model state/action capacity & 132 \\
Action representation & Arm relative, hand absolute; executed as absolute targets \\
State/action grouping & Named joint groups (arm, hand; both sides if bimanual) \\
State/action normalization & Min--max on training-set 1st/99th percentiles, clipped to $[-1, 1]$ \\
State dropout & 0.2 \\
Published pooled image processor & Random $230\times230$ crop resized to $256\times256$; no color jitter or rotation. Evaluation uses the corresponding center crop \\
\midrule
Optimizer & AdamW \\
AdamW $\beta_1, \beta_2$ & 0.9, 0.999 \\
Optimizer updates & 12,000 per standalone or pooled model \\
Global batch size & 32 \\
Micro-batch $\times$ accumulation & $2 \times 16$ on one GPU \\
Peak learning rate & $4 \times 10^{-5}$ \\
Learning-rate schedule & Linear warmup (5\% of updates, 600 steps) then cosine decay \\
Weight decay & $1 \times 10^{-5}$ \\
Gradient-norm clipping & 1.0 \\
Precision & BF16 \\
Gradient checkpointing & Disabled \\
Hardware & $1 \times$ A100-80GB \\
Training seed & 42 (one run per model; no seed sweep) \\
\midrule
Action prediction horizon & 40 steps \\
Actions executed per query & 8 \\
Inference denoising steps & 4 \\
Physics backend & PhysX CPU \\
Episode limit & Task's registered horizon \\
Evaluation checkpoint & Update 12,000 \\
Evaluation episodes & 50 per stage-level task, scene or long-horizon task \\
Evaluation seeds & Environment seeds $10001$--$10050$, one per episode; the same 50 seeds for every stage-level task, every generalization scene and the long-horizon protocol; the evaluation process is seeded with 10001 \\
\bottomrule
\end{tabular}
\end{subtable}
\end{table}
\FloatBarrier

\subsection{Dataset Statistics and Analysis}
\label{subsec:dataset_statistics_and_analysis}

\subsubsection{Hand Usage}
\label{subsubsec:hand_usage}
Table~\ref{tab:hand_usage_subtasks} gives the three statistics summarised in Figure~\ref{fig:hand_usage} for every stage-level task. 
Each stage-level task pools the frames of its 100 released episodes up to the first success. 
For the bimanual stage-level task (OpenPackage) the two hands are analyzed separately and averaged. 
The category rows are the values plotted in the main figure and quoted in Section~\ref{subsec:teleoperation_dataset}.
Category values weight stage-level tasks equally.

In Grasping and Insertion, the contact sits on the fingertip pads, led by the thumb pad (86\% and 82\% of contact frames) with the middle and index pads next; the ring pad joins in a third of the frames and the little finger and palm almost never. 
Twisting spreads contact over fewer frames and more links per finger, as the fingers release and re-contact the part. 
Pressing is the one category without thumb contact: the index pad and distal phalanx carry the button and key presses, while the middle and ring pads (37\% and 32\%) and the palm (25\%) come from holding the pipette.
Scooping and pouring wraps the whole thumb around the tool, with all three thumb links in contact for a third of the frames or more. 
Opening and closing shows the lightest and most evenly spread contact. 
Joint excursion is carried mainly by the knuckle flexion joints of the index, middle and ring finger, which move through 40--52\% of their range in grasping, twisting and scooping. 
Table~\ref{tab:hand_usage_subtasks} gives the timesteps to success, contact ratio and fingers in contact for every stage-level task; the category rows are the values plotted in Figure~\ref{fig:hand_usage}.

\begin{table}[t]
\centering\small
\caption{Timesteps to success, contact ratio and fingers in contact for every stage-level task, grouped by motion category. 
Timesteps: control steps (20\,Hz) to first success, median over the 100 episodes with the interquartile range in parentheses; category rows give the range of the stage-level task medians. Contact ratio: fraction of frames, up to first success, in which a finger or the palm touches the primary object of the stage-level task. 
Fingers: number of distinct fingers in contact, averaged over hand-contact frames. Stage-level task values pool their 100 episodes.
Category rows weight stage-level tasks equally.}
\label{tab:hand_usage_subtasks}
\setlength{\tabcolsep}{6pt}
\begin{tabular}{@{}lccc@{}}
\toprule
Stages & Timesteps & Contact ratio & Fingers \\
\midrule
\textbf{Opening and closing} & \textbf{60--162} & \textbf{0.52} & \textbf{2.3} \\
\quad OpenLunchBox & 99 (93--109) & 0.78 & 1.6 \\
\quad OpenCabinet & 110 (99--120) & 0.58 & 3.2 \\
\quad OpenDrawer & 103 (91--120) & 0.59 & 3.5 \\
\quad OpenFridge & 162 (134--183) & 0.63 & 3.6 \\
\quad CloseFridge & 60 (56--67) & 0.45 & 1.9 \\
\quad OpenLaptop & 101 (92--107) & 0.37 & 1.0 \\
\quad OpenPackage & 156 (134--208) & 0.23 & 1.1 \\
\midrule
\textbf{Grasping and placement} & \textbf{118--376} & \textbf{0.54} & \textbf{3.1} \\
\quad PickBottle & 208 (186--225) & 0.48 & 3.0 \\
\quad PickCharger & 201 (173--238) & 0.54 & 3.4 \\
\quad PickBolt & 118 (105--137) & 0.60 & 2.1 \\
\quad PlaceBread & 147 (131--173) & 0.59 & 3.7 \\
\quad ExtractApple & 248 (223--280) & 0.48 & 3.7 \\
\quad AppleToBlender & 249 (221--270) & 0.52 & 3.8 \\
\quad LayFlatPanel & 148 (130--184) & 0.69 & 2.4 \\
\quad PickDebris & 376 (350--412) & 0.40 & 2.5 \\
\midrule
\textbf{Insertion and assembly} & \textbf{155--248} & \textbf{0.68} & \textbf{2.7} \\
\quad InsertPlug & 224 (190--275) & 0.80 & 3.5 \\
\quad InsertRail & 248 (192--306) & 0.55 & 1.9 \\
\quad SeatNut & 155 (140--180) & 0.61 & 2.0 \\
\quad PlaceLid & 162 (140--182) & 0.77 & 3.5 \\
\midrule
\textbf{Twisting and screwing} & \textbf{41--310} & \textbf{0.47} & \textbf{2.1} \\
\quad UnscrewCap & 310 (250--356) & 0.60 & 2.5 \\
\quad TurnNut & 158 (133--209) & 0.48 & 2.1 \\
\quad FastenScrew & 41 (38--47) & 0.34 & 1.6 \\
\midrule
\textbf{Pressing} & \textbf{62--474} & \textbf{0.40} & \textbf{1.9} \\
\quad PressButton & 74 (64--86) & 0.13 & 1.0 \\
\quad ClickMouse & 62 (58--68) & 0.07 & 1.0 \\
\quad TypeKeys & 440 (398--480) & 0.22 & 1.0 \\
\quad AspirateLiquid & 326 (302--365) & 0.77 & 3.2 \\
\quad DispenseLiquid & 474 (432--521) & 0.80 & 3.4 \\
\midrule
\textbf{Scooping and pouring} & \textbf{145--332} & \textbf{0.71} & \textbf{3.1} \\
\quad ScoopTomato & 332 (288--369) & 0.81 & 3.0 \\
\quad PourBeads & 193 (168--213) & 0.65 & 4.4 \\
\quad DumpDustpan & 145 (138--154) & 0.67 & 2.0 \\
\bottomrule
\end{tabular}
\end{table}

\subsubsection{Episode Diversity}
\label{subsubsec:episode_diversity}

Table~\ref{tab:diversity_subtasks} measures how much the 100 episodes of a stage-level task differ in the motion of the hand. 
The palm position is obtained by forward kinematics from the recorded joint positions. 
Across stage-level tasks, the mean palm path length ranges from 0.35\,m to 2.04\,m, and the palm paths of two episodes are 3.3--10.2\,cm apart on average. 
The coefficient of variation of the time to success is 0.10--0.36.

\begin{table}[t]
\centering\small
\caption{Diversity of the released demonstrations, per stage-level task. Path length:
length of the palm path from the first frame to first success, mean over the 100
episodes. Pairwise distance: mean top-view distance between the palm paths of two
episodes after resampling both to 50 points in normalised time, averaged over all
episode pairs. Time CV: coefficient of variation of the timesteps to first
success. For the bimanual stage-level task (OpenPackage) the hand values are given as
left / right.}
\label{tab:diversity_subtasks}
\setlength{\tabcolsep}{6pt}
\begin{tabular}{@{}lccc@{}}
\toprule
Stages & Path length (m) & Pairwise distance (cm) & Time CV \\
\midrule
\textbf{Opening and closing} & & & \\
\quad OpenLunchBox & 0.53 & 3.9 & 0.13 \\
\quad OpenCabinet & 0.86 & 4.9 & 0.15 \\
\quad OpenDrawer & 0.60 & 3.9 & 0.20 \\
\quad OpenFridge & 1.11 & 6.9 & 0.21 \\
\quad CloseFridge & 1.38 & 8.7 & 0.13 \\
\quad OpenLaptop & 0.80 & 5.1 & 0.10 \\
\quad OpenPackage & 0.72 / 0.65 & 6.6 / 6.4 & 0.36 \\
\midrule
\textbf{Grasping and placement} & & & \\
\quad PickBottle & 1.40 & 7.9 & 0.14 \\
\quad PickCharger & 1.90 & 8.9 & 0.20 \\
\quad PickBolt & 0.76 & 7.2 & 0.18 \\
\quad PlaceBread & 0.99 & 5.5 & 0.24 \\
\quad ExtractApple & 1.75 & 7.0 & 0.17 \\
\quad AppleToBlender & 1.93 & 8.1 & 0.12 \\
\quad LayFlatPanel & 1.29 & 7.5 & 0.26 \\
\quad PickDebris & 1.57 & 6.2 & 0.12 \\
\midrule
\textbf{Insertion and assembly} & & & \\
\quad InsertPlug & 1.08 & 7.5 & 0.28 \\
\quad InsertRail & 1.13 & 10.2 & 0.29 \\
\quad SeatNut & 0.88 & 6.8 & 0.20 \\
\quad PlaceLid & 0.89 & 5.7 & 0.18 \\
\midrule
\textbf{Twisting and screwing} & & & \\
\quad UnscrewCap & 1.40 & 9.6 & 0.23 \\
\quad TurnNut & 0.88 & 7.1 & 0.34 \\
\quad FastenScrew & 0.50 & 5.1 & 0.29 \\
\midrule
\textbf{Pressing} & & & \\
\quad PressButton & 0.41 & 4.3 & 0.26 \\
\quad ClickMouse & 0.35 & 3.3 & 0.12 \\
\quad TypeKeys & 1.37 & 4.7 & 0.13 \\
\quad AspirateLiquid & 1.62 & 4.9 & 0.15 \\
\quad DispenseLiquid & 2.04 & 7.0 & 0.13 \\
\midrule
\textbf{Scooping and pouring} & & & \\
\quad ScoopTomato & 1.98 & 7.8 & 0.15 \\
\quad PourBeads & 1.13 & 8.5 & 0.17 \\
\quad DumpDustpan & 0.99 & 4.6 & 0.11 \\
\bottomrule
\end{tabular}
\end{table}

\subsubsection{Open-loop Replay}
\label{subsubsec:open-loop_replay}

Every accepted demonstration is replayed open loop. 
The recorded initial state is restored in a fresh environment, the recorded action sequence is executed step by step, and the episode counts as a replay success if the success condition becomes true at any step. 
The replay uses the same simulator build as the recording. 
Over the 30 stage-level tasks, 3222 of the 4112 accepted episodes (78\%) replay to success.
The released 100 episodes of each stage-level task are drawn from the replay-successful ones (Appendix~\ref{subsec:data_post-processing_validation_release}).

Table~\ref{tab:replay_subtasks} lists the replay success per stage-level task next to the three contact statistics used below. 
Replay success is high for Opening and Closing (98\%) and lowest for Insertion and Assembly (65\%) and Scooping and Pouring (74\%). 
Within the latter two categories it varies widely: PlaceLid and InsertPlug replay in 58\% and 62\% of episodes, ScoopTomato in 38\%, while DumpDustpan reaches 94\%.

Across the 30 stage-level tasks, replay success is negatively correlated with the contact ratio (Spearman $\rho=-0.58$, $p<0.001$), with the number of contact events per episode ($\rho=-0.51$, $p=0.004$) and with the number of fingers in contact ($\rho=-0.44$, $p=0.014$). 

Stage-level tasks involving more sustained or repeated contact therefore tend to have lower open-loop replay success. 
We observe trajectory deviations from the recorded rollouts shortly after replay begins, even though the recorded initial state and action sequence are restored. Together, these results are consistent with contact-rich motions being more sensitive to small state or simulator differences under open-loop replay.

\begin{table}[t]
\centering\small
\caption{Open-loop replay success of the accepted demonstrations, per stage-level task and
motion category, next to the three contact statistics it correlates with. Replay:
the recorded actions are replayed from the recorded initial state in a fresh
simulator instance; an episode counts as replayable if the success condition
becomes true at any step (replayable/accepted episodes in parentheses). Contact
ratio and fingers: as in Table~\ref{tab:hand_usage_subtasks}. Events: number of
separate hand-contact intervals with the primary object per episode, up to first
success, mean over episodes. Category rows weight stage-level tasks equally.}
\label{tab:replay_subtasks}
\setlength{\tabcolsep}{6pt}
\begin{tabular}{@{}lcccc@{}}
\toprule
Stages & Replay success & Contact ratio & Events & Fingers \\
\midrule
\textbf{Opening and closing} & \textbf{98\%} & \textbf{0.52} & \textbf{2.1} & \textbf{2.3} \\
\quad OpenLunchBox & 99\% (110/111) & 0.78 & 2.2 & 1.6 \\
\quad OpenCabinet & 99\% (100/101) & 0.58 & 1.9 & 3.2 \\
\quad OpenDrawer & 100\% (110/110) & 0.59 & 1.9 & 3.5 \\
\quad OpenFridge & 93\% (107/115) & 0.63 & 2.0 & 3.6 \\
\quad CloseFridge & 100\% (104/104) & 0.45 & 2.2 & 1.9 \\
\quad OpenLaptop & 100\% (109/109) & 0.37 & 2.7 & 1.0 \\
\quad OpenPackage & 97\% (102/105) & 0.23 & 2.0 & 1.1 \\
\midrule
\textbf{Grasping and placement} & \textbf{84\%} & \textbf{0.54} & \textbf{3.2} & \textbf{3.1} \\
\quad PickBottle & 83\% (110/132) & 0.48 & 3.3 & 3.0 \\
\quad PickCharger & 94\% (107/114) & 0.54 & 2.5 & 3.4 \\
\quad PickBolt & 74\% (107/145) & 0.60 & 2.6 & 2.1 \\
\quad PlaceBread & 84\% (102/122) & 0.59 & 2.4 & 3.7 \\
\quad ExtractApple & 92\% (104/113) & 0.48 & 3.6 & 3.7 \\
\quad AppleToBlender & 83\% (106/127) & 0.52 & 2.3 & 3.8 \\
\quad LayFlatPanel & 85\% (102/120) & 0.69 & 2.1 & 2.4 \\
\quad PickDebris & 74\% (101/136) & 0.40 & 6.4 & 2.5 \\
\midrule
\textbf{Insertion and assembly} & \textbf{65\%} & \textbf{0.68} & \textbf{3.5} & \textbf{2.7} \\
\quad InsertPlug & 62\% (101/163) & 0.80 & 2.2 & 3.5 \\
\quad InsertRail & 62\% (110/177) & 0.55 & 6.4 & 1.9 \\
\quad SeatNut & 80\% (105/132) & 0.61 & 2.8 & 2.0 \\
\quad PlaceLid & 58\% (116/199) & 0.77 & 2.7 & 3.5 \\
\midrule
\textbf{Twisting and screwing} & \textbf{87\%} & \textbf{0.47} & \textbf{4.7} & \textbf{2.1} \\
\quad UnscrewCap & 73\% (111/153) & 0.60 & 7.5 & 2.5 \\
\quad TurnNut & 96\% (118/123) & 0.48 & 4.9 & 2.1 \\
\quad FastenScrew & 94\% (119/127) & 0.34 & 1.8 & 1.6 \\
\midrule
\textbf{Pressing} & \textbf{83\%} & \textbf{0.40} & \textbf{3.1} & \textbf{1.9} \\
\quad PressButton & 100\% (100/100) & 0.13 & 1.3 & 1.0 \\
\quad ClickMouse & 99\% (114/115) & 0.07 & 1.1 & 1.0 \\
\quad TypeKeys & 97\% (111/115) & 0.22 & 5.9 & 1.0 \\
\quad AspirateLiquid & 81\% (101/125) & 0.77 & 3.6 & 3.2 \\
\quad DispenseLiquid & 39\% (108/280) & 0.80 & 3.8 & 3.4 \\
\midrule
\textbf{Scooping and pouring} & \textbf{74\%} & \textbf{0.71} & \textbf{2.0} & \textbf{3.1} \\
\quad ScoopTomato & 38\% (118/311) & 0.81 & 2.1 & 3.0 \\
\quad PourBeads & 89\% (107/120) & 0.65 & 2.0 & 4.4 \\
\quad DumpDustpan & 94\% (102/108) & 0.67 & 2.0 & 2.0 \\
\bottomrule
\end{tabular}
\end{table}

\subsection{Generalization Evaluation Scenes}
\label{appendix:generalization_scenes}
\begin{figure}[t]
    \centering
    \includegraphics[width=\linewidth]{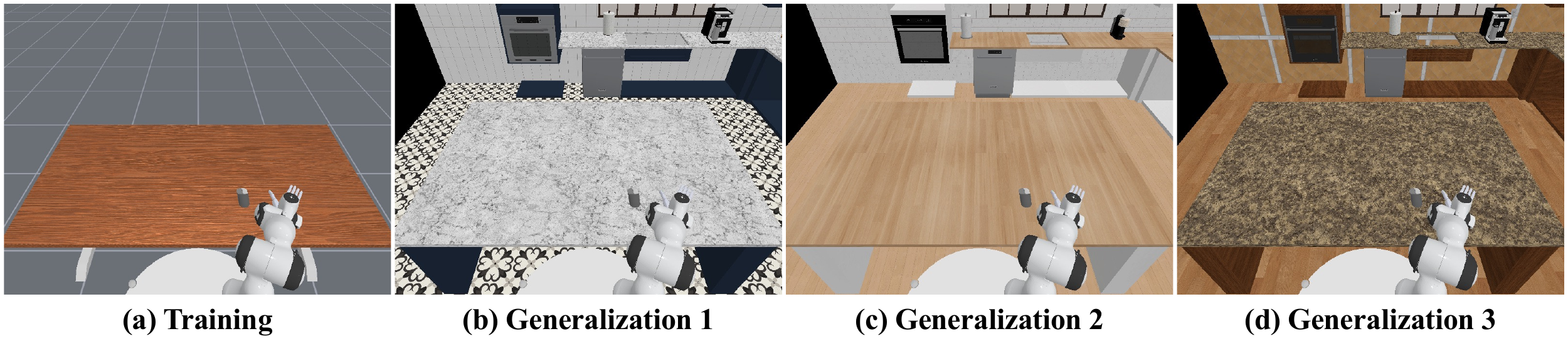}
    \caption{\textbf{Evaluation scenes for generalization.} Training and three held-out scene settings used to evaluate visual and environmental generalization across the 30 stage-level tasks. The scenes span three kitchen environments while preserving the task-specific objects and success conditions.}
    \label{fig:generalization_evaluation_scenes}
\end{figure}

The default training scene and the varied scenes for generalization evaluation are visualized in \Cref{fig:generalization_evaluation_scenes}.

\subsection{Long-horizon Performance}
\label{appendix:long_horizon_performance}

The evaluation protocol covers six baselines on the 10 long-horizon tasks.
For the single-task baselines, we reuse the learned stage-level policies without additional training and switch policies when the success condition of the current stage is met.
For both VLA baselines ($\pi_{0.5}$ and GR00T), we train one model per long-horizon task using demonstrations from all constituent stage-level tasks and switch only the language instruction between stages, keeping the same model checkpoint fixed.
The environment is initialized only at the beginning of the full task and is not reset between stages.
An episode succeeds only if all required stages are completed in order and the final success condition of the full task is satisfied within the evaluation horizon.

Table~\ref{tab:long-horizon-performance} reports end-to-end success rates and the current $\pi_{0.5}$-LoRA entries. For $\pi_{0.5}$, Meal preparation, Drawer and plug, Nut and bolt, and Table cleaning have independently verified zero success rates. The remaining task entries await independent final acceptance and are not measured policy-failure rates; its displayed ten-task average is not yet validated. These incomplete entries are excluded from comparative performance conclusions.

\begin{table*}[!t]
\centering
\caption{\textbf{Long-horizon performance.} End-to-end success rates (\%, $\uparrow$) on 10 long-horizon tasks. The evaluation protocol specifies 50 episodes per task.
An episode succeeds only if all required stages are completed in order within the evaluation horizon.
Both $\pi_{0.5}$ and GR00T use a single task-specific model trained on pooled stage demonstrations, with one fixed checkpoint per complete episode and instruction changes between stages.}
\label{tab:long-horizon-performance}
\fontsize{9.2}{11}\selectfont
\setlength{\tabcolsep}{6pt}
\renewcommand{\arraystretch}{1.18}

\begin{tabular*}{\textwidth}{@{\extracolsep{\fill}}l|cc|cc|cc@{}}
\toprule
\rowcolor{tablehead}
\textbf{Task}
& \multicolumn{2}{c|}{\textbf{Demo-guided RL}}
& \multicolumn{2}{c|}{\textbf{Imitation learning}}
& \multicolumn{2}{c}{\textbf{VLA models}} \\
\cmidrule(lr){2-3}\cmidrule(lr){4-5}\cmidrule(l){6-7}
\rowcolor{tablehead}
& \textbf{PPO} & \textbf{RLPD}
& \textbf{ACT} & \textbf{DP}
& \textbf{GR00T} & \shortstack{$\boldsymbol{\pi_{0.5}}$\\\textbf{(LoRA)}} \\
\midrule
\textbf{Meal preparation} & 0 & 0 & 0 & 0 & 0 & 0 \\
\textbf{Cabinet and pouring} & 0 & 0 & 0 & 0 & 0 & 0 \\
\textbf{Drawer and plug} & 0 & 0 & 0 & 0 & 0 & 0 \\
\textbf{Nut and bolt} & 0 & 0 & 0 & 0 & 0 & 0 \\
\textbf{Office desk} & 0 & 0 & 0 & 0 & 0 & 0 \\
\addlinespace[3pt]
\textbf{Furniture assembly} & 0 & 0 & 8 & 0 & 0 & 0 \\
\textbf{Pipette transfer} & 0 & 0 & 0 & 0 & 0 & 0 \\
\textbf{Juice preparation} & 0 & 0 & 0 & 0 & 0 & 0 \\
\textbf{Package opening} & 0 & 0 & 4 & 0 & 0 & 0 \\
\textbf{Table cleaning} & 0 & 0 & 0 & 0 & 0 & 0 \\
\midrule
\rowcolor{tablesummary}
\textbf{Average} & 0 & 0 & 1.2 & 0 & 0 & 0 \\
\bottomrule
\end{tabular*}
\end{table*}

% \begin{table*}[!t]
% \centering
% \caption{\textbf{Long-horizon performance.} End-to-end success rates (\%, $\uparrow$) on 10 long-horizon tasks, with 50 evaluation episodes per task.
% An episode succeeds only if all required stages are completed in order within the evaluation horizon.}
% \label{tab:long-horizon-performance}
% \fontsize{9.2}{11}\selectfont
% \setlength{\tabcolsep}{9pt}
% \renewcommand{\arraystretch}{1.18}

% \begin{tabular*}{\textwidth}{@{\extracolsep{\fill}}lccccc@{}}
% \toprule
% \rowcolor{tablehead}
% \textbf{Task} & \textbf{PPO} & \textbf{RLPD} & \textbf{ACT} & \textbf{DP} & \textbf{GR00T} \\
% \midrule
% \textbf{Meal preparation} & 0 & 0 & 0 & 0 & 0 \\
% \textbf{Cabinet and pouring} & 0 & 0 & 0 & 0 & 0 \\
% \textbf{Drawer and plug} & 0 & 0 & 0 & 0 & 0 \\
% \textbf{Nut and bolt} & 0 & 0 & 0 & 0 & 0 \\
% \textbf{Office desk} & 0 & 0 & 0 & 0 & 0 \\
% \addlinespace[3pt]
% \textbf{Furniture assembly} & 0 & 0 & 8 & 0 & 0 \\
% \textbf{Pipette transfer} & 0 & 0 & 0 & 0 & 0 \\
% \textbf{Juice preparation} & 0 & 0 & 0 & 0 & 0 \\
% \textbf{Package opening} & 0 & 0 & 4 & 0 & 0 \\
% \textbf{Table cleaning} & 0 & 0 & 0 & 0 & 0 \\
% \midrule
% \rowcolor{tablesummary}
% \textbf{Average} & 0 & 0 & 1.2 & 0 & 0 \\
% \bottomrule
% \end{tabular*}
% \end{table*}

\end{document}

%% file: math_commands.tex
\usepackage{amsmath,amsfonts,bm}

\def\eqref#1{equation~\ref{#1}}
\def\1{\bm{1}}

\DeclareMathAlphabet{\mathsfit}{\encodingdefault}{\sfdefault}{m}{sl}
\SetMathAlphabet{\mathsfit}{bold}{\encodingdefault}{\sfdefault}{bx}{n}

%% file: iclr2027_conference.bib
@misc{schulman2017ppo,
      title={Proximal Policy Optimization Algorithms}, 
      author={John Schulman and Filip Wolski and Prafulla Dhariwal and Alec Radford and Oleg Klimov},
      year={2017},
      eprint={1707.06347},
      archivePrefix={arXiv},
      primaryClass={cs.LG},
      url={https://arxiv.org/abs/1707.06347}, 
}

@InProceedings{ball2023rlpd,
  title = 	 {Efficient Online Reinforcement Learning with Offline Data},
  author =       {Ball, Philip J. and Smith, Laura and Kostrikov, Ilya and Levine, Sergey},
  booktitle = 	 {Proceedings of the 40th International Conference on Machine Learning},
  pages = 	 {1577--1594},
  year = 	 {2023},
  editor = 	 {Krause, Andreas and Brunskill, Emma and Cho, Kyunghyun and Engelhardt, Barbara and Sabato, Sivan and Scarlett, Jonathan},
  volume = 	 {202},
  series = 	 {Proceedings of Machine Learning Research},
  month = 	 {23--29 Jul},
  publisher =    {PMLR},
  url = 	 {https://proceedings.mlr.press/v202/ball23a.html}
}

@INPROCEEDINGS{zhao2023act, 
    AUTHOR    = {Tony Z. Zhao AND Vikash Kumar AND Sergey Levine AND Chelsea Finn}, 
    TITLE     = {{Learning Fine-Grained Bimanual Manipulation with Low-Cost Hardware}}, 
    BOOKTITLE = {Proceedings of Robotics: Science and Systems}, 
    YEAR      = {2023}, 
    ADDRESS   = {Daegu, Republic of Korea}, 
    MONTH     = {July}, 
    DOI       = {10.15607/RSS.2023.XIX.016} 
}

@inproceedings{chi2023diffusion,
	title={Diffusion Policy: Visuomotor Policy Learning via Action Diffusion},
	author={Chi, Cheng and Feng, Siyuan and Du, Yilun and Xu, Zhenjia and Cousineau, Eric and Burchfiel, Benjamin and Song, Shuran},
	booktitle={Proceedings of Robotics: Science and Systems (RSS)},
	year={2023}
}

@article{james2019rlbench,
  title={RLBench: The Robot Learning Benchmark \& Learning Environment},
  author={James, Stephen and Ma, Zicong and Rovick Arrojo, David and Davison, Andrew J.},
  journal={IEEE Robotics and Automation Letters},
  year={2020}
}

@inproceedings{gu2023maniskill2, author={Gu, Jiayuan and others}, title={ManiSkill2: A Unified Benchmark for Generalizable Manipulation Skills}, booktitle={ICLR}, year={2023}}

@InProceedings{yu2020metaworld,
  title = 	 {Meta-World: A Benchmark and Evaluation for Multi-Task and Meta Reinforcement Learning},
  author =       {Yu, Tianhe and Quillen, Deirdre and He, Zhanpeng and Julian, Ryan and Hausman, Karol and Finn, Chelsea and Levine, Sergey},
  booktitle = 	 {Proceedings of the Conference on Robot Learning},
  pages = 	 {1094--1100},
  year = 	 {2020},
  editor = 	 {Kaelbling, Leslie Pack and Kragic, Danica and Sugiura, Komei},
  volume = 	 {100},
  series = 	 {Proceedings of Machine Learning Research},
  month = 	 {30 Oct--01 Nov},
  publisher =    {PMLR},
  url = 	 {https://proceedings.mlr.press/v100/yu20a.html}
}

@InProceedings{Bao_2023_CVPR,
    author    = {Bao, Chen and Xu, Helin and Qin, Yuzhe and Wang, Xiaolong},
    title     = {DexArt: Benchmarking Generalizable Dexterous Manipulation With Articulated Objects},
    booktitle = {Proceedings of the IEEE/CVF Conference on Computer Vision and Pattern Recognition (CVPR)},
    month     = {June},
    year      = {2023},
    pages     = {21190-21200}
}

@INPROCEEDINGS{rajeswaran2018learningcomplexdexterousmanipulation, 
    AUTHOR    = {Aravind Rajeswaran AND Vikash Kumar AND Abhishek Gupta AND Giulia Vezzani AND John Schulman AND Emanuel Todorov AND Sergey Levine}, 
    TITLE     = {Learning Complex Dexterous Manipulation with Deep Reinforcement Learning and Demonstrations}, 
    BOOKTITLE = {Proceedings of Robotics: Science and Systems}, 
    YEAR      = {2018}, 
    ADDRESS   = {Pittsburgh, Pennsylvania}, 
    MONTH     = {June}, 
    DOI       = {10.15607/RSS.2018.XIV.049} 
}

@inproceedings{jiang2025dexmimicgenautomateddatageneration,
      title     = {DexMimicGen: Automated Data Generation for Bimanual Dexterous Manipulation via Imitation Learning},
      author    = {Jiang, Zhenyu and Xie, Yuqi and Lin, Kevin and Xu, Zhenjia and Wan, Weikang and Mandlekar, Ajay and Fan, Linxi and Zhu, Yuke},
      booktitle = {2025 IEEE International Conference on Robotics and Automation (ICRA)},
      year      = {2025}
}

@inproceedings{nasiriany2024robocasa,
  title={RoboCasa: Large-Scale Simulation of Everyday Tasks for Generalist Robots},
  author={Soroush Nasiriany and Abhiram Maddukuri and Lance Zhang and Adeet Parikh and Aaron Lo and Abhishek Joshi and Ajay Mandlekar and Yuke Zhu},
  booktitle={Robotics: Science and Systems (RSS)},
  year={2024}
}

@InProceedings{robotwin2025,
  author = {Mu, Yao and Chen, Tianxing and Chen, Zanxin and Peng, Shijia and Lan, Zhiqian and Gao, Zeyu and Liang, Zhixuan and Yu, Qiaojun and Zou, Yude and Xu, Mingkun and Lin, Lunkai and Xie, Zhiqiang and Ding, Mingyu and Luo, Ping},
  title = {RoboTwin: Dual-Arm Robot Benchmark with Generative Digital Twins},
  booktitle = {Proceedings of the Computer Vision and Pattern Recognition Conference (CVPR)},
  month = {June},
  year = {2025},
  pages = {27649-27660}
}

@article{robodojo2026,
  title={{RoboDojo}: A Unified Sim-and-Real Benchmark for Comprehensive Evaluation of Generalist Robot Manipulation Policies},
  author={Chen, Tianxing and Chen, Yue and Li, Zixuan and Tang, Junyuan and Su, Kailun and Wan, Weijie and Chen, Baijun and Lu, Haoran and Yan, Haowen and Su, Honghao and others},
  journal={arXiv preprint arXiv:2607.04434},
  year={2026}
}

@INPROCEEDINGS{black2024pi0, 
    AUTHOR    = {Kevin Black AND Noah Brown AND Danny Driess AND Adnan Esmail AND Michael Robert Equi AND Chelsea Finn AND Niccolo Fusai AND Lachy Groom AND Karol Hausman AND Brian Ichter AND Szymon Jakubczak AND Tim Jones AND Liyiming Ke AND Sergey Levine AND Adrian Li-Bell AND Mohith Mothukuri AND Suraj Nair AND Karl Pertsch AND Lucy Xiaoyang Shi AND Laura Smith AND James Tanner AND Quan Vuong AND Anna Walling AND Haohuan Wang AND Ury Zhilinsky}, 
    TITLE     = {{$\pi_0$: A Vision-Language-Action Flow Model for General Robot Control}}, 
    BOOKTITLE = {Proceedings of Robotics: Science and Systems}, 
    YEAR      = {2025}, 
    ADDRESS   = {LosAngeles, CA, USA}, 
    MONTH     = {June}, 
    DOI       = {10.15607/RSS.2025.XXI.010} 
}

@inproceedings{gr00tn1_2025,
  archivePrefix = {arxiv},
  eprint     = {2503.14734},
  title      = {{GR00T} {N1}: An Open Foundation Model for Generalist Humanoid Robots},
  author     = {NVIDIA and Johan Bjorck and Fernando Castañeda and Nikita Cherniadev and Xingye Da and Runyu Ding and Linxi "Jim" Fan and Yu Fang and Dieter Fox and Fengyuan Hu and Spencer Huang and Joel Jang and Zhenyu Jiang and Jan Kautz and Kaushil Kundalia and Lawrence Lao and Zhiqi Li and Zongyu Lin and Kevin Lin and Guilin Liu and Edith Llontop and Loic Magne and Ajay Mandlekar and Avnish Narayan and Soroush Nasiriany and Scott Reed and You Liang Tan and Guanzhi Wang and Zu Wang and Jing Wang and Qi Wang and Jiannan Xiang and Yuqi Xie and Yinzhen Xu and Zhenjia Xu and Seonghyeon Ye and Zhiding Yu and Ao Zhang and Hao Zhang and Yizhou Zhao and Ruijie Zheng and Yuke Zhu},
  month      = {March},
  year       = {2025},
  booktitle  = {ArXiv Preprint},
}

@misc{dexjoco2026,
      title={DexJoCo: A Benchmark and Toolkit for Task-Oriented Dexterous Manipulation on MuJoCo}, 
      author={Hanwen Wang and Weizhi Zhao and Xiangyu Wang and Siyuan Huang and He Lin and Boyuan Zheng and Rongtao Xu and Gang Wang and Yao Mu and He Wang and Lue Fan and Hongsheng Li and Zhaoxiang Zhang and Tieniu Tan},
      year={2026},
      eprint={2605.16257},
      archivePrefix={arXiv},
      primaryClass={cs.RO},
      url={https://arxiv.org/abs/2605.16257}, 
}

@misc{ye2026worldactionmodelszeroshot,
      title={World Action Models are Zero-shot Policies},
      author={Seonghyeon Ye and Yunhao Ge and Kaiyuan Zheng and Shenyuan Gao and Sihyun Yu and George Kurian and Suneel Indupuru and You Liang Tan and Chuning Zhu and Jiannan Xiang and Ayaan Malik and Kyungmin Lee and William Liang and Nadun Ranawaka and Jiasheng Gu and Yinzhen Xu and Guanzhi Wang and Fengyuan Hu and Avnish Narayan and Johan Bjorck and Jing Wang and Gwanghyun Kim and Dantong Niu and Ruijie Zheng and Yuqi Xie and Jimmy Wu and Qi Wang and Ryan Julian and Danfei Xu and Yilun Du and Yevgen Chebotar and Scott Reed and Jan Kautz and Yuke Zhu and Linxi "Jim" Fan and Joel Jang},
      year={2026},
      eprint={2602.15922},
      archivePrefix={arXiv},
      primaryClass={cs.RO},
      url={https://arxiv.org/abs/2602.15922},
}

@InProceedings{Xiang_2020_CVPR,
    author = {Xiang, Fanbo and Qin, Yuzhe and Mo, Kaichun and Xia, Yikuan and Zhu, Hao and Liu, Fangchen and Liu, Minghua and Jiang, Hanxiao and Yuan, Yifu and Wang, He and Yi, Li and Chang, Angel X. and Guibas, Leonidas J. and Su, Hao},
    title = {SAPIEN: A SimulAted Part-Based Interactive ENvironment},
    booktitle = {Proceedings of the IEEE/CVF Conference on Computer Vision and Pattern Recognition (CVPR)},
    month = {June},
    year = {2020}
}

@inproceedings{qin2023anyteleop,
  title     = {AnyTeleop: A General Vision-Based Dexterous Robot Arm-Hand Teleoperation System},
  author    = {Qin, Yuzhe and Yang, Wei and Huang, Binghao and Van Wyk, Karl and Su, Hao and Wang, Xiaolong and Chao, Yu-Wei and Fox, Dieter},
  booktitle = {Robotics: Science and Systems},
  year      = {2023}
}

@misc{sharpa2026manussdk,
  title        = {{Manus SDK Integration for SharpaWave}},
  author       = {{Sharpa Robotics}},
  year         = {2026},
  howpublished = {\url{https://github.com/sharpa-robotics/sharpa-manus-sdk}},
  note         = {Commit b87bff5, accessed 2026-09-22}
}

@inproceedings{physicalintelligence2025pi05,
  title = {{$\pi_{0.5}$}: A Vision-Language-Action Model with Open-World Generalization},
  author = {Black, Kevin and Brown, Noah and Darpinian, James and Dhabalia, Karan and Driess, Danny and Esmail, Adnan and Equi, Michael Robert and Finn, Chelsea and Fusai, Niccolo and Galliker, Manuel Y. and Ghosh, Dibya and Groom, Lachy and Hausman, Karol and Ichter, Brian and Jakubczak, Szymon and Jones, Tim and Ke, Liyiming and LeBlanc, Devin and Levine, Sergey and Li-Bell, Adrian and Mothukuri, Mohith and Nair, Suraj and Pertsch, Karl and Ren, Allen Z. and Shi, Lucy Xiaoyang and Smith, Laura and Springenberg, Jost Tobias and Stachowicz, Kyle and Tanner, James and Vuong, Quan and Walke, Homer and Walling, Anna and Wang, Haohuan and Yu, Lili and Zhilinsky, Ury},
  booktitle = {Proceedings of The 9th Conference on Robot Learning},
  pages = {17--40},
  year = {2025},
  volume = {305},
  series = {Proceedings of Machine Learning Research},
  publisher = {PMLR},
  url = {https://proceedings.mlr.press/v305/black25a.html}
}

@inproceedings{hu2022lora,
  title = {{LoRA}: Low-Rank Adaptation of Large Language Models},
  author = {Hu, Edward J. and Shen, Yelong and Wallis, Phillip and Allen-Zhu, Zeyuan and Li, Yuanzhi and Wang, Shean and Wang, Lu and Chen, Weizhu},
  booktitle = {International Conference on Learning Representations},
  year = {2022},
  url = {https://openreview.net/forum?id=nZeVKeeFYf9}
}

@misc{cadene2024lerobot,
    author = {Cadene, Remi and Alibert, Simon and Soare, Alexander and Gallouedec, Quentin and Zouitine, Adil and Palma, Steven and Kooijmans, Pepijn and Aractingi, Michel and Shukor, Mustafa and Aubakirova, Dana and Russi, Martino and Capuano, Francesco and Pascal, Caroline and Choghari, Jade and Meftah, Khalil and Ellerbach, Maxime and Moss, Jess and Wolf, Thomas},
    title = {LeRobot: State-of-the-art Machine Learning for Real-World Robotics in Pytorch},
    howpublished = "\url{https://github.com/huggingface/lerobot}",
    year = {2024}
}

@inproceedings{todorov2012mujoco,
  title={MuJoCo: A physics engine for model-based control},
  author={Todorov, Emanuel and Erez, Tom and Tassa, Yuval},
  booktitle={2012 IEEE/RSJ International Conference on Intelligent Robots and Systems},
  pages={5026--5033},
  year={2012},
  organization={IEEE},
  doi={10.1109/IROS.2012.6386109}
}
